\documentclass[a4paper,fleqn]{cas-sc}

\usepackage[
  backend=biber,
  style=authoryear,
  natbib=true,
  maxcitenames=2,
  maxbibnames=99,
  giveninits=true,
  uniquename=init
]{biblatex}
\usepackage{siunitx}
\usepackage{subcaption}
\usepackage{enumitem}
\usepackage{placeins}
\usepackage{tikz}
\usetikzlibrary{arrows.meta,calc,positioning,shapes.geometric}

\usepackage[nameinlink,noabbrev]{cleveref}
\usepackage{csquotes}

\newcommand{\stdfigurewidth}{0.98\linewidth}
\begin{document}
\let\WriteBookmarks\relax
% Suppress the auto-rendered "ORCID(s):" footnote placeholder from cas-sc
% (no author ORCIDs declared via \orcidauthor; leaving \printorcid active
%  prints an empty "ORCID(s):" line on the title page).
\let\printorcid\relax
\renewcommand{\topfraction}{0.9}
\renewcommand{\bottomfraction}{0.8}
\renewcommand{\textfraction}{0.07}
\renewcommand{\floatpagefraction}{0.8}
\setcounter{topnumber}{3}
\setcounter{bottomnumber}{2}
\setcounter{totalnumber}{5}

\shorttitle{Transfer Learning for Fragility Adaptation}
\shortauthors{Saeednejad and Padgett}

\title[mode=title]{Bridging Data Gaps in Structural Fragility Modeling through Transfer Learning: Methodology and Case Studies}

\author[1]{Narges Saeednejad}
\credit{Conceptualization, Methodology, Coding, Investigation, Data collection and curation, Visualization, Writing}

\author[1,2]{Jamie Ellen Padgett}
\ead{jamie.padgett@rice.edu}
\cormark[1]
\credit{Conceptualization, Methodology, Resources, Supervision, Funding acquisition, Review \& editing}

\cortext[1]{Corresponding author}

\affiliation[1]{organization={Department of Civil and Environmental Engineering, Rice University},
            city={Houston},
            state={TX},
            country={USA}}

\affiliation[2]{organization={Ken Kennedy Institute, Rice University},
            city={Houston},
            state={TX},
            country={USA}}

\begin{abstract}
Fragility functions underpin catastrophe modeling, risk analysis, and resilience assessments by relating hazard and structural variables to the probability of structural damage or failure.
In practice, fragility model libraries are highly uneven: data-rich regions and well-studied hazards enable detailed model development, while many domains remain data-scarce, heterogeneous, and subject to distributional mismatch between available training data and target portfolios.
This paper presents a methodology-centered transfer learning framework for fragility adaptation under domain shift, class imbalance, and scarce target labels while preserving engineering interpretability and supporting decision-making under uncertainty.
Four transfer learning strategies (instance-based, parameter-based, hierarchical Bayesian, and multi-source) are demonstrated through three complementary case studies: (i) instance-based transfer learning via importance weighting, demonstrated on coastal bridge fragility using Hurricane Katrina observations; (ii) parameter-based transfer learning together with hierarchical Bayesian transfer learning, enabling partial pooling across strata and posterior uncertainty quantification, demonstrated on residential building fragility using Hurricane Ian observations; and (iii) multi-source transfer learning that fuses multiple analytical fragility models with learned source weights and regularized target-domain adaptation, demonstrated on seismic bridge fragility using observations from the 2001 Nisqually earthquake.
Across these case studies, direct transfer of source models (i.e. using existing state-of-the-art models) fails under domain shift and severe class imbalance, while targeted adaptation substantially improves failure detection and predictive stability in low-data regimes. These findings highlight the need for systematic guidance on diagnostics, strategy selection, and uncertainty reporting when developing and adapting fragility models.
\end{abstract}

% Highlights block omitted from the arXiv version (the Elsevier
% cas-sc class renders \begin{highlights}...\end{highlights} on a
% separate front page, which is not needed for the arXiv posting).


\maketitle

% ========================================
% SECTION 1: INTRODUCTION
% ========================================

\section{Introduction}
\label{sec:introduction}

Fragility functions are essential tools for probabilistic risk analysis because they quantify the conditional probability of reaching or exceeding a damage state as a function of hazard intensity and structural parameters. As such, they are foundational to performance-based engineering, regional risk assessment, catastrophe modeling, and resilience-informed decision-making across the disaster lifecycle, from mitigation and design to emergency response and recovery planning \parencite{ellingwood_fragility_2004,lallemant_statistical_2015}. In coastal storm and seismic settings, where infrastructure systems are exposed to low-frequency but high-consequence hazards, the need for reliable fragility models is especially pressing. This need is further amplified by growing hazard exposure in coastal communities, increasing concentration of assets in vulnerable regions, and the cascading societal consequences that follow infrastructure disruption \parencite{crossett2004population,neumann_future_2015,webster_changes_2005}.

Classical analytical approaches remain important because they provide interpretable relationships between structural demand, capacity, and damage \parencite{porter_creating_2007}. At the same time, physics-based and simulation-driven approaches have enabled richer representation of nonlinear structural behavior, hazard uncertainty, and system-level response \parencite{ataei2015fragility, baker2015efficient, vanDeLindt2009performance, karamlou2015computation}. Statistical procedures for deriving fragility curves from analytical, empirical, or hybrid datasets have also become increasingly rigorous \parencite{porter_creating_2007,lallemant_statistical_2015,zentner_fragility_2017}. For structural systems with multiple interacting components, fragility methods have increasingly been extended to capture component correlations and system dependencies through multicomponent and system-level formulations, including coupled demand-based and mixture-based representations, or Bayesian networks \parencite{bandini2022seismic,gehl2016development}. Bayesian formulations have further expanded the fragility toolbox by supporting calibration, model updating, and uncertainty quantification as new evidence becomes available.\parencite{kennedy2001bayesian,li2013bayesian,zhang2022conjugate,yan2024hierarchical,mishra2017hurricane,lei2021seismic}. Furthermore, approaches for representing the fragility of heterogenous structural portfolios \parencite{rincon2024fragility} have evolved from the use of archetypes (or representative structures) \parencite{hazus2024eq}, to increasingly refined sub-class representations \parencite{chen2025secondgen}, to parameterized fragilities intended to afford tailored estimates across a range of inventory characteristics \parencite{balomenos2020parameterized}, to the potential for structure specific fragilities derived directly from surrogates \parencite{lee2025systematic}.  

Despite this methodological progress, the practical use of fragility models remains constrained by highly uneven data and model availability across hazards, regions, structural systems, and design configurations. Consequently, important gaps persist in fragility libraries for many hazard–structure combinations, particularly in understudied systems and new application settings. In these cases, practitioners are often left to either apply models outside their intended context or undertake the long and tedious process of rebuilding fragility models from scratch for each new setting. This limitation motivates strategies that can transfer knowledge from existing fragility models and damage data to fill such gaps more rapidly and efficiently. In other cases, best-estimate models are calibrated using post-event observations and reconnaissance data \parencite{cremen2019benchmarking}, insurance records \parencite{wing2020new}, remote-sensing products \parencite{forte2025flash}, or synthetic simulation datasets generated through analytical or numerical modeling \parencite{baker2015efficient}. These calibration efforts can improve regional realism, but they are often limited by sparse observations, inconsistent inventory attributes, and missing data \parencite{little2019missing}. For example, post-disaster bridge damage datasets are typically small and strongly imbalanced, with failure cases remaining rare even after major events \parencite{padgett2008bridge,ranf_damage_nodate}. Similar challenges arise in broader fragility and vulnerability modeling, where limited real-world observations often force analysts to rely on simplified assumptions, numerical surrogates, or archetype-based approximations \parencite{nofal2020minimal,reis2022methodology,harirchian2024utilizing}. As a result, fragility models developed in data-rich source settings are frequently transferred informally to data-scarce target settings without sufficient validation of whether the underlying distributions remain compatible.

This limitation is fundamentally a \textit{domain shift} problem. From a statistical and machine-learning perspective, direct transfer becomes unreliable when the source and target domains differ in their feature distributions and, more generally, in their joint predictor--outcome structure. Under such mismatch, a model that is well calibrated in one region or inventory may yield biased predictions in another. This issue has been widely studied in the transfer-learning and domain-adaptation literature, where standard classifiers are known to degrade when training and testing data are not drawn from the same distribution \parencite{pan_survey_2010,zhuang_comprehensive_2021,kouw_introduction_2019}. In particular, covariate-shift theory formalizes the need to reweight source samples by the density ratio $p_T(x)/p_S(x)$ to reduce target-domain bias, although estimating these weights robustly remains difficult when target data are scarce \parencite{sugiyama2007covariate,kouw_introduction_2019}. For fragility modeling, this means that physically informed base models are not automatically transferable: they must be checked for compatibility and, when necessary, adapted to the target domain rather than deployed unchanged.

Transfer learning provides a principled framework to address this challenge by reusing information learned in data-rich source domains to improve inference in related but data-scarce target domains \parencite{pan_survey_2010,zhuang_comprehensive_2021}. Broadly speaking, transfer can occur through reweighting or selecting source instances, adapting parameters of an existing model, transferring informative priors within a Bayesian framework, or combining information from multiple source models. Related ideas have already shown promise in civil-engineering applications beyond fragility modeling. Transfer learning has been used for image-based structural damage recognition under limited labeled data \parencite{gao2018deep}, structural health monitoring and anomaly detection across different structures \parencite{ierimonti2021transfer,pan2023transfer,bao2023deep}, and hurricane damage classification from imagery \parencite{calton2022using}. These studies collectively demonstrate the value of transferring learned knowledge when labeled target data are limited or when source and target domains are related but not identical. However, most of this literature focuses on classification, detection, or condition identification tasks rather than on the adaptation of fragility functions themselves.

Accordingly, an important gap remains in fragility modeling: how to adapt existing fragility models when target-region data are scarce, class imbalance is severe, and the source and target domains are not statistically aligned. Comparatively less attention has been paid to diagnosing when direct transfer is likely to fail, selecting an adaptation strategy that matches the available data regime, and producing updated fragility relationships that preserve engineering interpretability while also quantifying uncertainty. This gap is especially important for safety-critical applications, where false confidence in an unadapted model can be more harmful than admitting limited information.

To address this need, this paper develops a transfer learning framework for fragility model adaptation under data scarcity, domain shift, and class imbalance. The proposed framework formulates adaptation as a systematic knowledge-transfer task. The study presents and demonstrates four complementary transfer learning strategies (instance-based, parameter-based, hierarchical Bayesian, and multi-source) through three case studies: (i) instance-based transfer learning for fragility adaptation under covariate shift, illustrated through a coastal bridge hurricane fragility case study using data from Hurricane Katrina in 2005; (ii) parameter-based transfer learning combined with hierarchical Bayesian transfer learning for residential building fragility under sparse and heterogeneous observations, illustrated using residential damage data from Hurricane Ian in 2022; and (iii) multi-source transfer learning for seismic bridge fragility adaptation when multiple analytical source models are available but none transfers reliably in isolation, illustrated using bridge damage observations from the 2001 Nisqually earthquake. Through these case studies, the paper provides methodological guidance and empirical evidence demonstrating that transfer learning can bridge persistent data gaps in fragility modeling while preserving the interpretability and uncertainty quantification required for risk-informed engineering decision-making.

% ========================================
% SECTION 2: METHODOLOGY
% ========================================

\section{Methodology}
\label{sec:methodology}

\subsection{Transfer learning framework for fragility model adaptation}
\label{sec:method_framework}

Rather than developing fragility models independently for every data-scarce region, this study formulates the problem as one of model adaptation: given sufficient labeled data in a source domain, or a base fragility model trained on source-domain data, the goal is to develop a reliable target-domain model by leveraging the limited damage observations available in the target region.Let the source domain be $\mathcal{D}_S = \{(\mathbf{x}_i^S, y_i^S)\}_{i=1}^{n_S}$ and the target domain be $\mathcal{D}_T = \{(\mathbf{x}_j^T, y_j^T)\}_{j=1}^{n_T}$, where each labeled observation consists of a predictor vector and its corresponding observed damage outcome. Here, $\mathbf{x}$ denotes the vector of hazard intensities and structural parameters, and $y$ denotes the observed damage outcome, represented either as a binary failure indicator or as an assigned damage state. For multi-state fragility formulations, let $DS_k$ denote the $k$th damage-state threshold, so that fragility is written as $P(DS \geq DS_k \mid \mathbf{x})$. The fundamental challenge arises when the joint distribution differs across domains, i.e., $p_S(\mathbf{x},y) \neq p_T(\mathbf{x},y)$, so that direct deployment of a source fragility model may produce biased risk estimates and degraded failure detection in the target region \parencite{pan_survey_2010,zhuang_comprehensive_2021}. Throughout the adaptation process, two objectives must be maintained simultaneously: preserving engineering interpretability and improving predictive reliability and stability under data scarcity, severe class imbalance, and domain shift.

The transfer learning approaches proposed for fragility adaptation and implemented in this study fall into four complementary categories, summarized in Table~\ref{tab:tl_strategies}. In \textit{instance-based transfer learning}, source samples are selected or reweighted to align with the target feature distribution, providing a practical approximation of importance weighting under covariate shift; this strategy is most effective when a labeled source dataset is available (i.e. original data used in base fragility derivation) but only limited labeled target observations can be obtained. In \textit{parameter-based transfer learning}, a trusted source fragility functional form is retained while model coefficients are recalibrated using limited target data, and target-specific covariates may be incorporated through regularized model augmentation to capture local vulnerability characteristics without overfitting. \textit{Bayesian transfer learning} encodes source-domain knowledge through informative priors, or hyperpriors in hierarchical formulations, and updates them with target observations, enabling partial pooling of information across strata and rigorous uncertainty quantification through posterior predictive distributions \parencite{karbalayghareh2018optimal,shwartz2022pretrain}. Finally, in \textit{multi-source transfer learning}, multiple existing fragility models are combined through learned source weights, while target-specific adjustments are regularized to reduce over-reliance on any single source and to mitigate negative transfer \parencite{sun2011twostage}.

\begin{table}[width=\linewidth,cols=4,pos=htbp]
\caption{Summary of transfer learning strategies for fragility model adaptation.}
\label{tab:tl_strategies}
\small
\renewcommand{\arraystretch}{1.25}
\begin{tabular*}{\tblwidth}{@{}p{2.8cm}p{3.8cm}p{3.2cm}p{2.5cm}@{}}
\toprule
Strategy & Mechanism & Key assumptions & Case study \\
\midrule
Instance-based &
Reweight or select source samples by similarity to the target feature distribution &
Labeled source dataset available; covariate shift is the primary barrier to transfer &
CS~I: coastal bridges (Hurricane Katrina) \\
\addlinespace
Parameter-based &
Retain source functional form; recalibrate coefficients with regularized fine-tuning on target data &
Trusted, physically grounded source model; target data sufficient for local coefficient adjustment &
CS~II: residential buildings (Hurricane Ian) \\
\addlinespace
Bayesian &
Encode source knowledge as informative priors; update with target data via hierarchical models &
Prior/hyperprior structure can represent source knowledge; subgroup partial pooling beneficial &
CS~II: residential buildings (Hurricane Ian) \\
\addlinespace
Multi-source &
Fuse multiple source models through learned weights with regularized target-specific deviations &
Multiple candidate source models available; no single source expected to transfer reliably alone &
CS~III: seismic bridges (Nisqually earthquake) \\
\bottomrule
\end{tabular*}
\end{table}

\subsection{Recommended workflow for fragility adaptation}
\label{sec:workflow}

\begin{figure}[pos=htbp]
  \centering
  \includegraphics[width=0.95\linewidth]{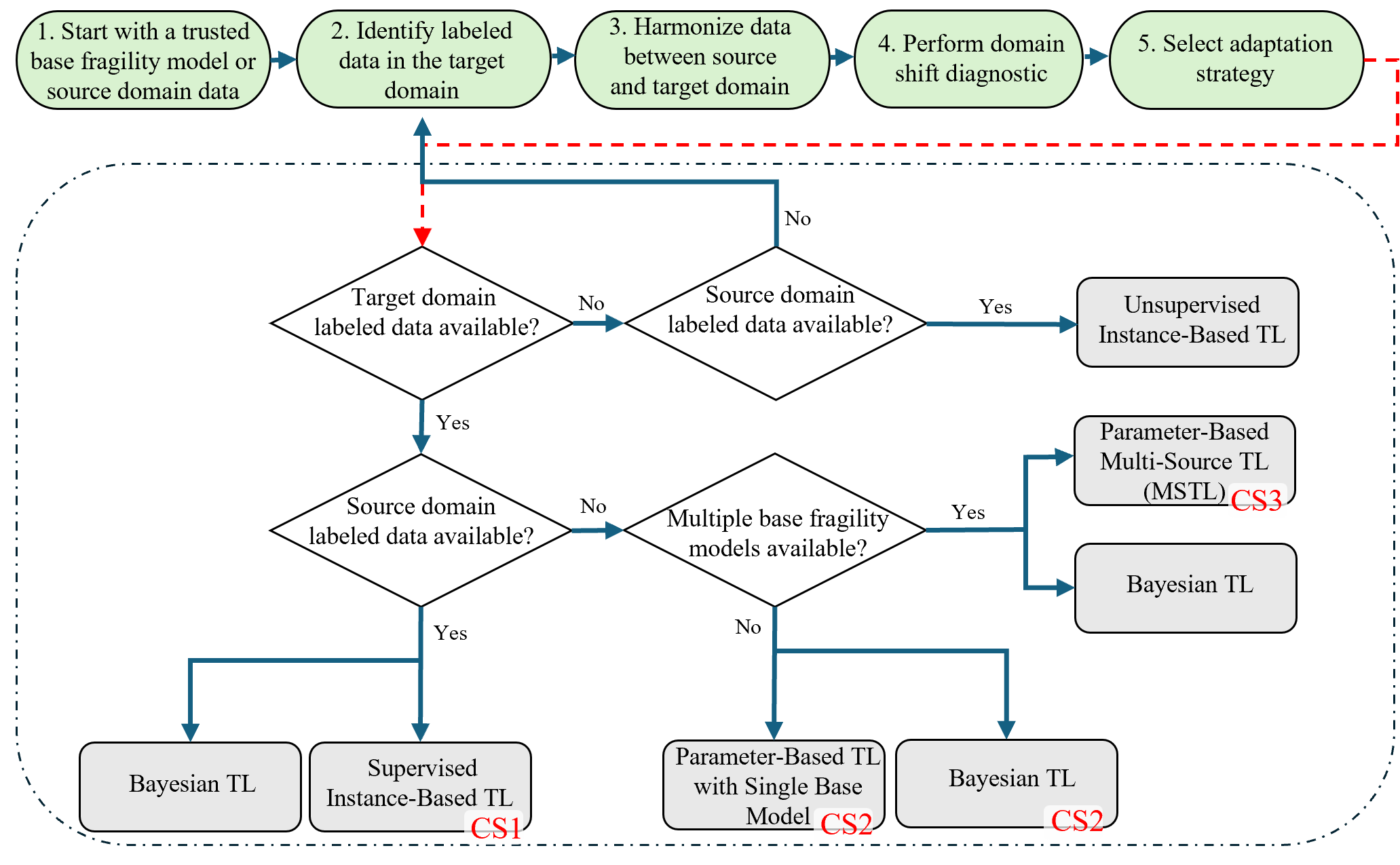}
  \caption{Workflow and decision tree for selecting a transfer learning strategy in fragility model adaptation. The case studies illustrating each strategy are indicated as CS1, CS2, and CS3.}
  \label{fig:framework_summary}
\end{figure}

Building on the framework above, a five-stage workflow is proposed for adapting fragility models to new regions under limited damage observations and potential domain shift. The workflow synthesizes the methodological strategies demonstrated in the case studies and is designed to integrate into standard fragility modeling pipelines (whether analytical, simulation-based, empirical, expert-judgement-driven, or hybrid) while preserving engineering interpretability. Figure~\ref{fig:framework_summary} summarizes the five stages and the decision tree that maps the available data regime to an adaptation strategy; the case studies illustrating each strategy are indicated in the figure as CS1, CS2, and CS3.

The first stage is \textit{start with a trusted base fragility model or source-domain data}. The workflow opens by assembling the source-domain knowledge available for adaptation. This may take the form of a trusted base fragility model \(\mathcal{M}_S(\mathbf{x};\boldsymbol{\theta}_S)\) (for example, a fitted logistic or lognormal fragility relationship from a prior study) or, when no such model is available, a raw labeled source dataset \(\mathcal{D}_S\) from a related region or component class. Both routes anchor the subsequent adaptation in pre-existing fragility knowledge and reduce the labeled-data requirement in the target region.

The second stage is \textit{identify labeled data in the target domain}. The labeled target-domain dataset \(\mathcal{D}_T = \{(\mathbf{x}_i^T, y_i^T)\}_{i=1}^{n_T}\) is then characterized. Both the sample size \(n_T\) and the class composition matter: small samples and an imbalanced ratio between failure (\(n_1\)) and non-failure (\(n_0\)) observations, with \(n_1 \ll n_0\) in many post-event datasets, restrict which adaptation strategies are viable.When labeled target data are entirely unavailable, adaptation must rely on unsupervised mechanisms that use labeled source data and unlabeled target data (predictor features without failure observations), as indicated in the decision tree in Figure~\ref{fig:framework_summary} indicates.

The third stage is \textit{harmonize data between source and target domain}. Once the source and target inputs are identified, the predictor vectors \(\mathbf{x}^S\) and \(\mathbf{x}^T\) are mapped to a unified feature space that preserves the meaning of hazard, structural, and contextual variables, and the damage states \(y^S\) and \(y^T\) are made comparable. Because fragility relationships from prior studies may take different functional forms, such as logistic, lognormal cumulative, or other parametric shapes, this harmonization step ensures that downstream adaptation operates on comparable representations.

The fourth stage is \textit{perform domain shift diagnostic}. Once the data are harmonized, the statistical compatibility between the source and target domains is assessed. Transfer may be unreliable when the joint distributions differ,
\[
p_S(\mathbf{x},y) \neq p_T(\mathbf{x},y),
\]
or, more specifically, when the feature distributions \(p_S(\mathbf{x})\) and \(p_T(\mathbf{x})\) are not aligned. Such mismatch may arise from geographic, structural, or hazard-related differences and can significantly degrade model transferability. Recommended diagnostics include comparisons of marginal feature distributions, correlation structures, and multivariate distance measures; kernel-based metrics such as Maximum Mean Discrepancy (MMD) are particularly useful for quantifying divergence between the source and target feature spaces \parencite{zhuang_comprehensive_2021}. The diagnostic identifies the covariates driving the mismatch and informs whether adaptation mechanisms such as instance-based approach, parameter adaptation or source-model fusion are required.

The fifth and final stage is \textit{select the adaptation strategy}. With the available source knowledge, target labels, harmonized inputs, and domain-shift evidence in hand, this stage chooses how the target-domain counterpart \(\mathcal{M}_T(\mathbf{x};\boldsymbol{\theta}_T)\) is constructed. The decision tree in Figure~\ref{fig:framework_summary} traces the strategy based on three binary checks: whether labeled target data are available, whether labeled source data are available, and whether multiple base fragility models are at hand. When both labeled target and labeled source data are available, instance-based transfer learning (selecting or reweighting source rows that are closest to the target distribution) becomes feasible, with a Bayesian variant providing principled uncertainty (Case Study~I, CS1). When labeled target data are available with no labeled source data and only a single trusted base model exists, parameter-based fine-tuning updates \(\boldsymbol{\theta}_S\) toward \(\boldsymbol{\theta}_T\) under regularization, with a hierarchical Bayesian formulation when uncertainty quantification is required (Case Study~II, CS2). When multiple credible base fragility models are available, parameter-based multi-source transfer learning (MSTL) fuses them through learned source weights, with a Bayesian counterpart available for uncertainty-aware predictions (Case Study~III, CS3). When labeled target data are entirely unavailable, unsupervised instance-based transfer learning remains an option, with the understanding that target-region validation is limited until target labels are observed.

% ========================================
% SECTION 3: CASE STUDIES
% ========================================

\section{Case studies for fragility adaptation}
\label{sec:case_studies}

The workflow presented in Section~\ref{sec:workflow} is now demonstrated through three case studies, each targeting a common fragility-adaptation challenge: distributional mismatch and covariate shift between source and target feature spaces, limited labeled target observations with heterogeneous subgroups, and source-model heterogeneity and negative-transfer risk. For each case study, the general methodological formulation is presented first, followed by an illustrative implementation using real infrastructure data and post-event observations.

% -------------------------------------------------
\subsection{Case Study I: Instance-based transfer learning under covariate shift}
\label{sec:case1}

\subsubsection{Methodology}
\label{sec:case1_methodology}

\paragraph{Problem setting.}

This case study addresses the transfer setting in which a labeled source dataset
$\mathcal{D}_S = \{(\mathbf{x}_i^S, y_i^S)\}_{i=1}^{n_S}$ is available together with
only limited labeled observations from the target domain
$\mathcal{D}_T = \{(\mathbf{x}_j^T, y_j^T)\}_{j=1}^{n_T}$, with $n_T \ll n_S$.
This setting is common in fragility practice when the labeled records used to fit
an existing fragility model can be reused, while only sparse post-event damage
observations are accessible in a related target domain. Direct deployment of the
source fragility model under covariate shift is biased in proportion to the
density-ratio mismatch $p_T(\mathbf{x})/p_S(\mathbf{x})$
\parencite{sugiyama2007covariate}. The instance-based transfer
learning (TL) strategy proposed here layers three complementary
mechanisms on top of standard similarity weighting
\parencite{pan_survey_2010,zhuang_comprehensive_2021}, so that the
contribution of each source row is governed simultaneously by
similarity to the target, by distribution-level covariate-shift
correction, and by label-level consistency with the target's
decision boundary.

\paragraph{Source model.}

The source model of \textcite{balomenos2020parameterized}, fit to the
Houston--Galveston simulation dataset $\mathcal{D}_S$, is the logistic
regression
\begin{equation}
P_f(\mathbf{x}) = \frac{1}{1 + \exp[-\ell(\mathbf{x})]},
\label{eq:case1_source_logistic}
\end{equation}
with polynomial logit
\begin{equation}
\begin{aligned}
\ell(\mathbf{x}) ={}& -14.08 - 6.15\,Z_c + 4.46\,H_{\max} + 0.02\,L_{sp} + 19.22\,|s_p| + 221.17\,|m_{sp}| \\
&{}+ 0.97\,(Z_c H_{\max}) + 25.39\,(Z_c\,|m_{sp}|) - 39.37\,(H_{\max}\,|m_{sp}|) - 0.67\,(s_p m_{sp}) \\
&{}- 174.17\,(|s_p|\,|m_{sp}|) - 0.96\,Z_c^{2} - 0.45\,H_{\max}^{2} - 12.30\,|s_p|^{2} \\
&{}- 9.20\times 10^{-5}\,(L_{sp} W_{sp} h) - 2.38\times 10^{-4}\,(L_{sp} W_{gd} n_{gd}),
\end{aligned}
\label{eq:case1_balomenos}
\end{equation}
where the predictors are defined in Table~\ref{tab:case1_source_vars}.
Eq.~\eqref{eq:case1_balomenos} is deployed without modification as the
\emph{direct-transfer} baseline on the Hurricane Katrina target inventory.

\begin{table}[width=.9\linewidth,cols=2,pos=htbp]
\caption{Case Study I: predictors of the source model of \textcite{balomenos2020parameterized} used in Eq.~\eqref{eq:case1_balomenos}.}
\label{tab:case1_source_vars}
\begin{tabular*}{\tblwidth}{@{}LL@{}}
\toprule
Variable & Description \\
\midrule
$Z_c$       & Relative surge elevation at bridge deck (m) \\
$H_{\max}$  & Maximum wave height at bridge location (m) \\
$L_{sp}$    & Span length (m) \\
$W_{sp}$    & Bridge deck width (m) \\
$h$         & Deck slab thickness (m) \\
$s_p$       & Wave spatial variability ($-$) \\
$m_{sp}$    & Span slope ($-$) \\
$W_{gd}$    & Girder weight (kN/m) \\
$n_{gd}$    & Number of girders ($-$) \\
\bottomrule
\end{tabular*}
\end{table}

\paragraph{Similarity-feature reweighting.}

Source samples are first reweighted by their radial basis function
(RBF) similarity to the target centroid in a standardized similarity
space $\mathbf{z}$,
\begin{equation}
w_i^{(\mathrm{rbf})} = \exp\!\left(-\frac{\|\mathbf{z}_i^S - \bar{\mathbf{z}}_T\|^2}{2\sigma^2}\right),
\label{eq:case1_rbf}
\end{equation}
with bandwidth $\sigma$ from the median-distance heuristic. The matching
is performed class-conditionally, so that source failures are
compared against target failures and non-failures against
non-failures; this preserves the target's class structure when the
two domains have different class priors.

\paragraph{Density-ratio importance weighting.}

A logistic domain classifier $g_\eta(\mathbf{x})$ is fit on
$\mathcal{D}_S \cup \mathcal{D}_T$ with class-balanced weights that
neutralize the imbalance $n_S \gg n_T$, and its class posterior is
used to estimate the covariate-shift importance weight
\begin{equation}
w_i^{(\mathrm{dr})} = \frac{\hat{P}(\mathrm{target}\mid \mathbf{x}_i^S)}{1 - \hat{P}(\mathrm{target}\mid \mathbf{x}_i^S)},
\label{eq:case1_dr}
\end{equation}
which is the standard density-ratio estimator of $p_T/p_S$ \parencite{sugiyama2007covariate}. The raw ratio is normalized so its
median equals unity and capped symmetrically to prevent any single
outlier from dominating the pool.

\paragraph{Pseudo-label consistency filter.}

A target-only logistic regression is fit on $\mathcal{D}_T$ and used
to assign a pseudo-label probability $\hat{p}_i$ to each candidate
source row. Source rows are retained only when their pseudo-label
agrees with the recorded label $y_i^S$ with confidence above a margin
$\delta$:
\begin{equation}
\mathcal{K} = \Bigl\{ i : (y_i^S = 1 \wedge \hat{p}_i \ge 0.5 + \delta) \vee (y_i^S = 0 \wedge \hat{p}_i \le 0.5 - \delta) \Bigr\},
\label{eq:case1_plf}
\end{equation}
with $\delta = 0.20$ in the present study. The pseudo-label model is
deliberately fit at the target's natural class prior rather than with
class-balanced weights; otherwise the decision boundary would shift
toward the minority (failure) class and reject most borderline source
non-failures, breaking the symmetry that the margin in
Eq.~\eqref{eq:case1_plf} is intended to enforce. A safeguard returns
the unfiltered pool when fewer than 30\% of candidate rows would
survive, so that the filter cannot silently collapse the pool when
the target sample is too small to define a reliable boundary.

\paragraph{Target fragility specification and pooled fitting.}

To define a target-domain fragility model, the predictor vector is
augmented with two physics-informed features related to wave-induced
deck uplift
\parencite{douglass2006impact,bradner2011experimental,aashto2008guide}.
The first feature is the absolute freeboard,
$\phi_1(\mathbf{x}) = |Z_{cb}-H_{\max}|$, which measures how close the
maximum wave height is to the bridge deck soffit. Smaller values of this
term indicate that the wave crest is closer to the deck level, a
condition associated with stronger uplift demand. The second feature is
the wave-to-slab thickness ratio, $\phi_2(\mathbf{x}) = H_{\max}/h$,
which represents the wave height relative to the deck slab thickness.

Using these features, the target-domain fragility logit is written as
\begin{equation}
\ell_T(\mathbf{x}) =
\beta_0 + \beta_1 H_{\max}
+ \beta_2 |Z_{cb}-H_{\max}|
+ \beta_3 H_{\max}/h .
\label{eq:case1_logit}
\end{equation}
The logit $\ell_T(\mathbf{x})$ is then converted to a failure probability
through the sigmoid, function:
\begin{equation}
P_f = \sigma(\ell_T)
= \frac{1}{1+\exp[-\ell_T(\mathbf{x})]} .
\label{eq:case1_fragility}
\end{equation}
This transformation maps the logit value to a probability between 0 and
1, where larger values of $\ell_T(\mathbf{x})$ correspond to higher
predicted probability of failure.

The model is fit using both the target training data and the selected
source samples. The selected source samples are included to augment the
limited target data, but their influence is controlled by transfer
weights so that source rows more consistent with the target domain
contribute more strongly to the fit. The model parameters are estimated
by minimizing the regularized weighted empirical risk
\begin{equation}
\min_{\theta}
\left[
\sum_{j \in \mathcal{D}_T^{(-k)}}
\omega_{y_j^T}\,
\ell\!\left(y_j^T, f_\theta(\mathbf{x}_j^T;\boldsymbol{\phi})\right)
+
\sum_{i \in \tilde{\mathcal{D}}_S}
w_i\,\omega_{y_i^S}\,
\ell\!\left(y_i^S, f_\theta(\mathbf{x}_i^S;\boldsymbol{\phi})\right)
+
\lambda \|\theta\|_2^2
\right],
\label{eq:case1_objective}
\end{equation}
where $\tilde{\mathcal{D}}_S$ is the selected source pool, $w_i$ is the
transfer weight assigned to source sample $i$, $\omega_c$ is the class
weight for class $c$, and $\lambda$ controls the strength of $L_2$
regularization. The class weights reduce the effect of imbalance between
failure and non-failure observations, while the regularization term helps
prevent overfitting. Model evaluation is performed using stratified
five-fold cross-validation, with the class weights and regularization
strength selected within the training fold.

\subsubsection{Illustrative example: Hurricane fragility adaptation for steel-girder bridges}
\label{sec:case1_example}
\paragraph{Data, imputation, and domain shift.}

\begin{figure}[pos=htbp]
  \centering
  \includegraphics[width=\stdfigurewidth]{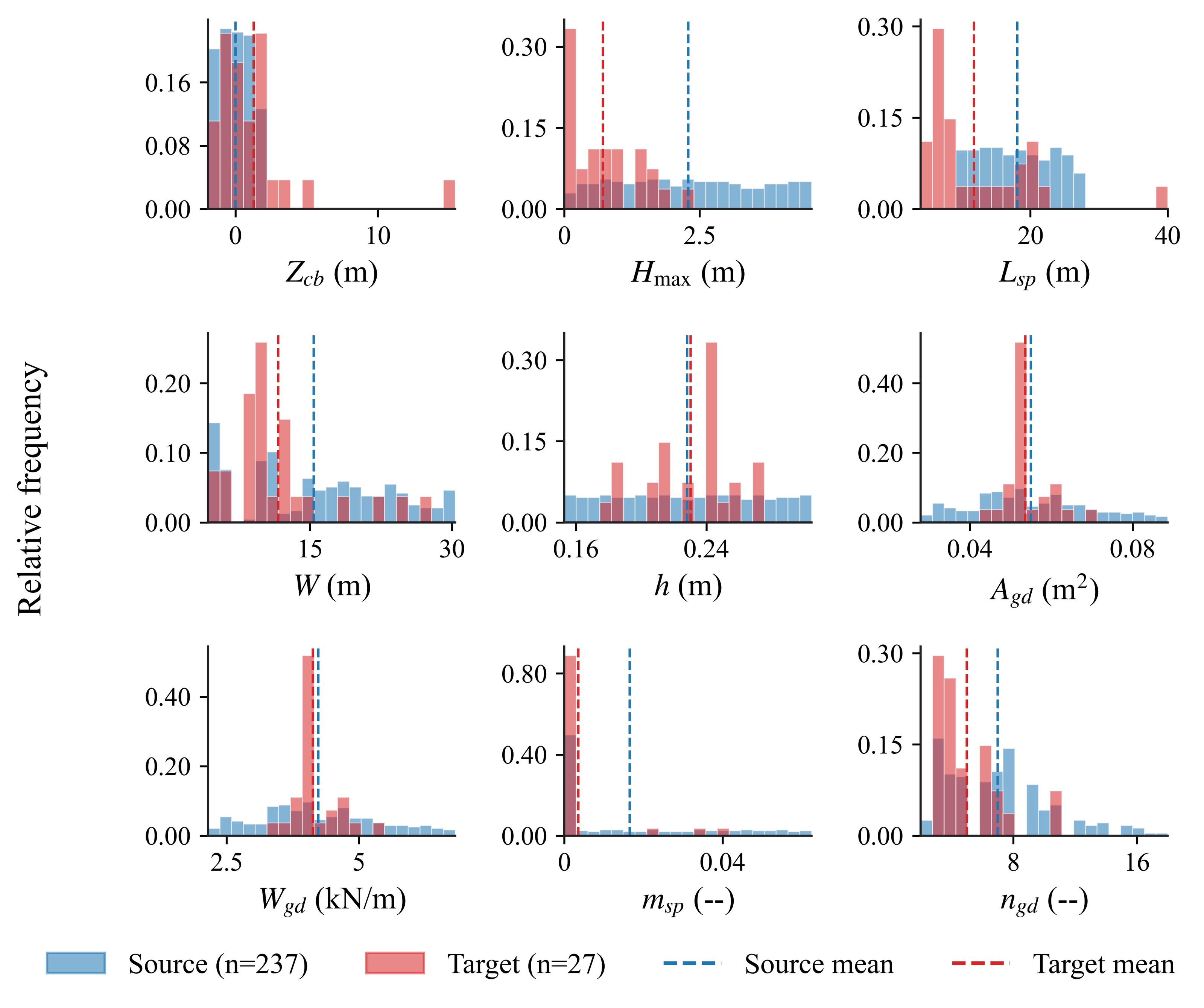}
  \caption{Case Study I: source ($n_S = 237$) versus target ($n_T = 29$ raw inventory; $n_T = 27$ after removing two surge-elevation outliers) marginal feature distributions.}
  \label{fig:case1_dist_compare}
\end{figure}

In the present application, the full target dataset comprises 29 observed steel-girder spans affected by Hurricane katrina
across the Gulf Coast \parencite{padgett2008bridge}, paired with
hindcast hazard intensities from the ADCIRC/SWAN coupled
storm-surge--wave model \parencite{kaiser2023adcirc}. For each bridge, the lowest span, representing the most hydraulically vulnerable section of the crossing, is used as the reference location for extracting deck elevation, surge, and wave-height values. Deck elevations are obtained from USGS 3DEP LiDAR, while surge and wave-height samples are taken at the corresponding span location rather than at the structure-level NBI centroid, which may be several kilometers away from the actual failure-prone deck region for long causeway crossings.
Where span-level structural attributes were unavailable in the
post-event inventory, missing values were imputed using
similarity-based information from the simulation dataset to obtain a
complete target feature matrix. Two records with inconsistent relative
surge-elevation conventions were then removed prior to model fitting,
reducing the target dataset from $n_T = 29$ (raw inventory) to
$n_T = 27$ steel-girder spans used for model fitting, including five
observed failures and an empirical failure rate of approximately 19\%. The corresponding source pool contains
$n_S = 237$ simulated steel-girder spans, with an approximately 54\%
failure rate. This difference indicates a substantial class-prior shift
between the source and target domains, in addition to the marginal
feature-distribution mismatch shown in Figure~\ref{fig:case1_dist_compare}.

Similarity diagnostics were used to identify a subset of structural
features with adequate overlap between the source and target domains
for source-sample selection. The selected similarity space includes slab
thickness $h$, girder cross-section $A_{gd}$, and girder unit weight
$W_{gd}$.

\begin{figure}[pos=htbp]
  \centering
  \includegraphics[width=\stdfigurewidth]{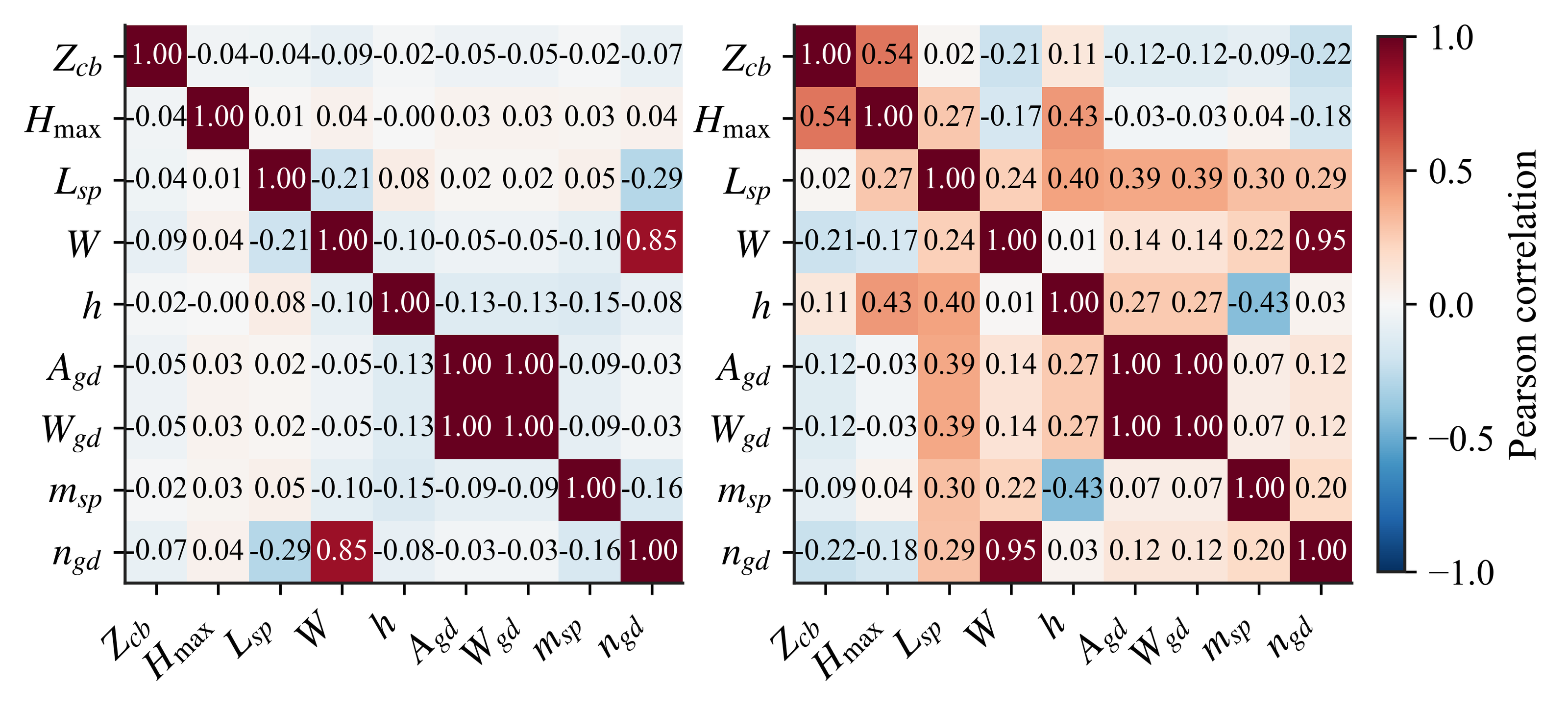}
  \par\vspace{1pt}
  {\small\sffamily\makebox[\stdfigurewidth][c]{%
    \makebox[\dimexpr\stdfigurewidth/2\relax][c]{\textbf{(a)} Source domain}%
    \makebox[\dimexpr\stdfigurewidth/2\relax][c]{\textbf{(b)} Target domain}%
  }}
  \caption{Case Study I: Pearson correlation matrices of the predictor set in the source and target inventories.}
  \label{fig:case1_corr_compare}
\end{figure}

Figure~\ref{fig:case1_dist_compare} shows clear marginal differences
between the source and target domains across several hazard and
structural variables. The source and target inventories also differ in
their multivariate dependence structure, as shown by the Pearson
correlation matrices in Figure~\ref{fig:case1_corr_compare}. In the
source domain, geometric variables such as $L_{sp}$ and $W$ are only
weakly correlated with structural variables such as $h$, $A_{gd}$, and
$W_{gd}$, whereas the target inventory exhibits stronger
geometric--structural coupling. These differences indicate that the
source and target domains are not aligned in either marginal feature
distributions or joint predictor relationships. Direct reuse of the
source fragility model is therefore unlikely to provide reliable target
predictions without adaptation, motivating the instance-based
reweighting and filtering strategy used in this case study.

\paragraph{Pool construction and domain-shift diagnostic.}

\begin{figure}[pos=htbp]
  \centering
  \includegraphics[width=0.93\linewidth]{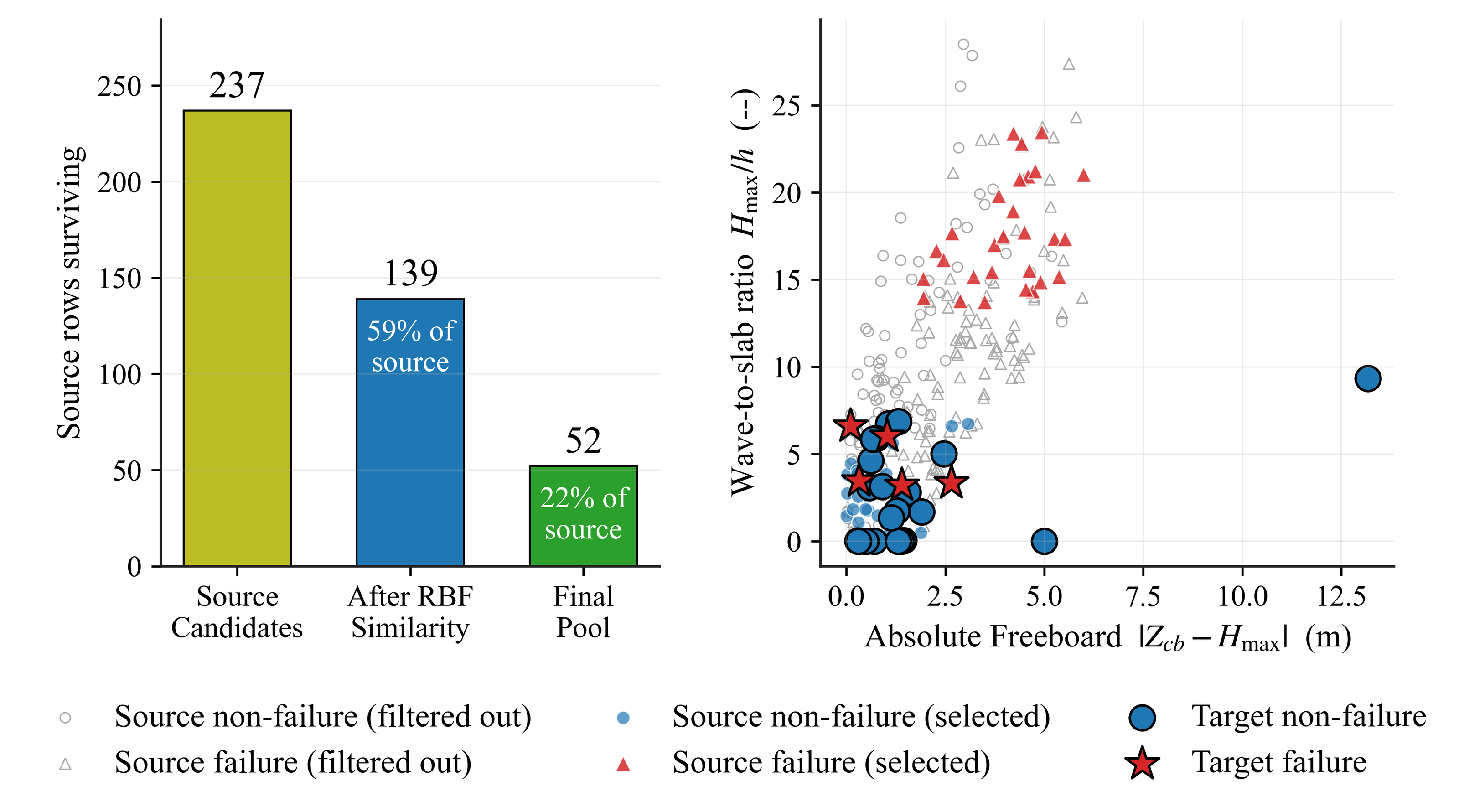}\\[-2pt]
  \noindent\begin{minipage}[t]{0.475\linewidth}
    {\footnotesize (a) Number of steel-girder source rows retained after each selection stage; the final pool applies the pseudo-label consistency filter ($\delta = 0.20$) and density-ratio weighting.\par}
  \end{minipage}\hfill
  \begin{minipage}[t]{0.475\linewidth}
    {\footnotesize (b) Selected source pool, 52 of 237 rows, shown in the engineered feature space $(|Z_{cb} - H_{\max}|,\; H_{\max}/h)$ used to fit the fragility model.\par}
  \end{minipage}
  \caption{Case Study I: instance-based TL source pool.}
  \label{fig:case1_pool}
\end{figure}

Figure~\ref{fig:case1_pool}(a) summarizes the source-sample selection
process. Starting from 237 candidate steel-girder source spans, the
class-conditional RBF similarity step retains 139 samples, and the
pseudo-label consistency filter further reduces the final transferable
source pool to 52 samples. The density-ratio weights of
Eq.~\eqref{eq:case1_dr} are then applied to adjust the contribution of
the retained source samples during model fitting, rather than to remove
additional samples.

The final source pool contains 28 failure cases and 24 non-failure
cases, providing a more balanced and target-relevant training set than
the original source catalogue. Importantly, the pseudo-label model is
used only as a filtering mechanism to identify source samples that are
consistent with the target-domain decision structure; it does not
replace the original simulation-derived source labels used for training.
Figure~\ref{fig:case1_pool}(b) shows the retained and excluded source
samples in the engineered feature space used for fragility fitting.

\paragraph{Quantitative alignment via MMD.}

The effectiveness of the source-selection pipeline is further evaluated
using Maximum Mean Discrepancy (MMD), a kernel-based measure of
distributional discrepancy commonly used in transfer learning
\parencite{gretton2012kernel,zhuang_comprehensive_2021}. The unbiased
$\mathrm{MMD}^2$ is computed between the source and target samples in
the structural similarity space $\{h, A_{gd}, W_{gd}\}$ using a
Gaussian RBF kernel with bandwidth selected by the median-distance
heuristic. Before selection, the full source pool yields
$\mathrm{MMD}^2 = 0.063$ relative to the target data. After applying
the selection pipeline and the combined RBF--density-ratio weights, the
weighted $\mathrm{MMD}^2$ decreases to approximately zero, indicating
substantial improvement in source--target alignment within the
similarity space. The same trend is observed within each outcome class,
with $\mathrm{MMD}^2$ decreasing from $0.070$ to approximately zero for
non-failures and from $0.012$ to approximately zero for failures.

An ablation study further confirms the importance of the pseudo-label
consistency filter. When the filter is removed and only the RBF and
density-ratio steps are retained, the selected pool increases to
$n_S = 139$, but predictive performance deteriorates substantially
(AUC $= 0.52$, Brier $= 0.29$, log-loss $= 1.38$). These results indicate that
distributional similarity alone is insufficient in this small-sample
setting; label-level consistency is also needed to avoid transferring
source samples that are feature-similar but inconsistent with the
target-domain decision structure.

\paragraph{Models compared and predictive performance.}

Three models are evaluated on the cleaned target dataset using
stratified five-fold cross-validation. The first, denoted
\emph{direct transfer}, is the source fragility model of
\textcite{balomenos2020parameterized} deployed without adaptation.
The second, the \emph{target-only model}, is a logistic
regression on the engineered feature set of
Eq.~\eqref{eq:case1_fragility} trained on $\mathcal{D}_T$ alone. The
third, the \emph{target model with selected source samples}, is the
proposed instance-based TL formulation of
Eq.~\eqref{eq:case1_objective} using the same engineered features as
the target-only model. Holding the feature representation
constant between the second and third models isolates the
contribution of the selected source samples from the contribution of
the feature engineering itself.

\begin{table}[width=.9\linewidth,cols=5,pos=t]
\caption{Case Study I: probabilistic metrics on the
cleaned Katrina target, aggregated over
five-fold stratified cross-validation with 1000-resample bootstrap
95\% confidence intervals. Best per metric in \textbf{bold}.}
\label{tab:case1_metrics}
\begin{tabular*}{\tblwidth}{@{}LCCCC@{}}
\toprule
Method & AUC ($\uparrow$) & AP ($\uparrow$) & Brier ($\downarrow$) & log-loss ($\downarrow$) \\
\midrule
Direct transfer & $0.514$ {\scriptsize $[0.18, 0.83]$} & $0.322$ & $0.340$ & $1.825$ \\
Target model without adaptation & $0.655$ {\scriptsize $[0.41, 0.88]$} & $0.334$ & $0.226$ & $0.644$ \\
\textbf{Target model with selected source samples} & $\mathbf{0.725}$ {\scriptsize $[\mathbf{0.53}, 0.91]$} & $\mathbf{0.383}$ & $\mathbf{0.190}$ & $\mathbf{0.608}$ \\
\bottomrule
\end{tabular*}
\end{table}

Table~\ref{tab:case1_metrics} compares the probabilistic performance
of the direct-transfer baseline, the target-only model, and the
proposed target model trained with selected source samples. Direct
transfer performs near chance, with AUC $=0.514$ and a wide bootstrap
confidence interval whose lower bound falls well below 0.5. This poor
performance is consistent with the substantial source--target mismatch
identified earlier, including both feature-distribution differences and
class-prior shift. The target-only model improves performance
(AUC $=0.655$), but its confidence interval still overlaps the chance
level, indicating limited stability under the small target sample size.
The proposed model with selected source samples achieves the strongest
overall performance, with the highest AUC
($0.725$), lowest Brier score
($0.190$), and lowest log-loss ($0.608$). Relative to the target-only
baseline, the selected-source model reduces the Brier score by
approximately 16\% and the log-loss by approximately 6\%; relative to
direct transfer, the reductions are approximately 44\% and 67\%,
respectively.

The improvement obtained from selected source samples can be interpreted
as a form of targeted sample-size augmentation. The target-only model in
Eq.~\eqref{eq:case1_fragility} contains four fitted parameters but is
trained with only five observed failure cases, making coefficient
estimation unstable. Although the $L_2$ penalty in
Eq.~\eqref{eq:case1_objective} helps control overfitting, it also
shrinks the fitted coefficients toward zero when target data alone are
used. By contrast, adding 52 source samples that are aligned with the
target domain provides additional information for estimating the
engineered-feature effects while retaining regularization. This allows
the adapted model to recover a stronger and more physically meaningful
response surface than the target-only fit. Table~\ref{tab:case1_fitted_coeffs} reports the fitted coefficients of the
two models: the selected-source fit recovers coefficient magnitudes that
are roughly an order of magnitude larger than the target-only fit (for
example, $\beta_2$ shifts from $-0.0643$ to $-0.634$), consistent with
the regularization-induced shrinkage of the target-only baseline.

\begin{table}[width=.9\linewidth,cols=3,pos=htbp]
\caption{Case Study I: fitted coefficients of the target-domain fragility logit (Eq.~\eqref{eq:case1_fragility}) for the target-only and selected-source models.}
\label{tab:case1_fitted_coeffs}
\begin{tabular*}{\tblwidth}{@{}LCC@{}}
\toprule
Coefficient & Target-Only & Target with Selected Source Samples \\
\midrule
$\beta_0$ & $-0.496$ & $-5.27$ \\
$\beta_1$ & $+0.311$ & $+2.11$ \\
$\beta_2$ & $-0.0643$ & $-0.634$ \\
$\beta_3$ & $+0.0902$ & $+0.539$ \\
\bottomrule
\end{tabular*}
\end{table}

\begin{figure}[pos=htbp]
  \centering
  \includegraphics[width=\linewidth]{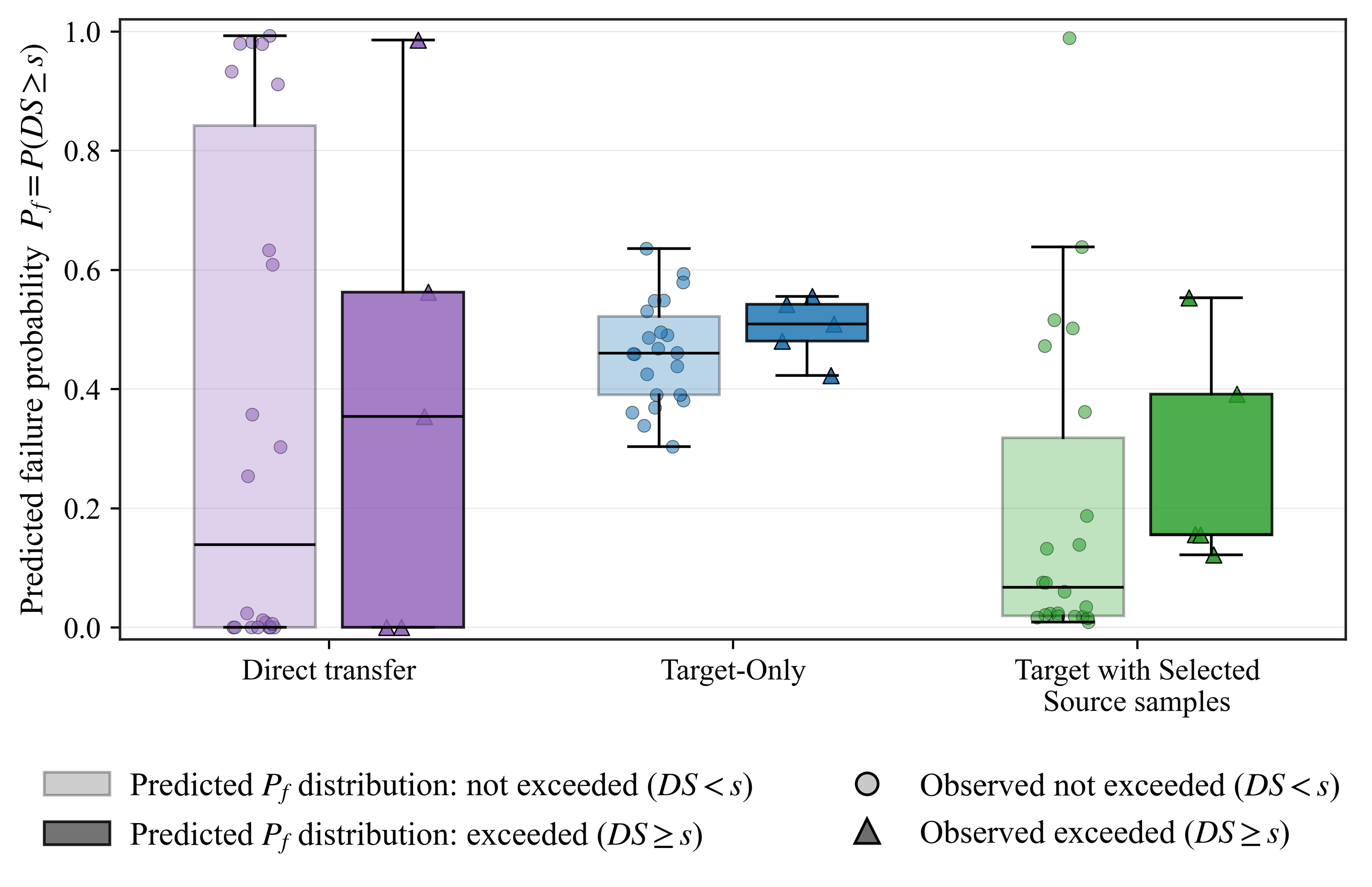}
  \caption{Case Study I: distribution of predicted failure probability
  $\hat{P}_f$, stratified by
  observed outcome and model. Boxplots summarize the predicted
  probability distributions for non-failure and failure observations
  for the three models reported in Table~\ref{tab:case1_metrics}.}
  \label{fig:case1_pfdist}
\end{figure}

Figure~\ref{fig:case1_pfdist} further illustrates the differences among
the three models by showing the distributions of
predicted failure probability, stratified by observed outcome. For the
target-only model, the predicted probabilities for failures and
non-failures are concentrated in a narrow range around the
$\hat{P}_f=0.5$ decision threshold, indicating weak separation between
classes. Direct transfer produces highly dispersed probabilities for
both classes, reflecting unstable extrapolation of the source fragility
surface to the target inventory. In contrast, the selected-source model
shifts non-failure predictions toward lower probabilities while moving
failure predictions upward, improving both class separation and
probabilistic calibration. Because the target dataset contains only a small number of observed
failures, conventional classification metrics may provide an incomplete
assessment of model performance. The evaluation therefore emphasizes
threshold-free and probability-sensitive metrics, including AUC, Brier
score, and log-loss, which quantify discriminative ability, calibration,
and predicted probability quality under severe data scarcity.

\paragraph{Discussion.}

The results of Case Study I highlight the importance of targeted
adaptation under simultaneous domain shift and data scarcity. Direct
transfer of the source fragility model performs poorly for the cleaned
target inventory of 27 spans with five observed failures, yielding the
lowest performance across all metrics in Table~\ref{tab:case1_metrics}.
Although the target-only model benefits from the proposed wave-uplift
feature representation, its predictions remain weakly separated because
the limited number of failure observations provides insufficient
information for stable parameter estimation.

In contrast, the target model trained with selected source samples
provides a more reliable probabilistic fragility estimate. The 52
retained source rows, selected through similarity weighting,
density-ratio correction, and label-consistency filtering, act as a
target-relevant augmentation of the limited training set. This adapted
model achieves the strongest overall performance, including the only AUC
confidence interval that excludes chance-level discrimination, while
reducing the Brier score and log-loss by 16\% and 6\%, respectively,
relative to the target-only baseline. Relative to direct transfer, the
corresponding reductions are 44\% and 67\%. These results demonstrate
that source information can improve target-domain fragility estimation
when it is selectively transferred rather than reused without adaptation.

% -------------------------------------------------
\subsection{Case Study II: Parameter-based and hierarchical Bayesian transfer learning}
\label{sec:case2}

\subsubsection{Methodology}
\label{sec:case2_methodology}

\paragraph{Problem setting.}

This case study addresses a different transfer setting from Case Study I. There, a labeled source dataset was available, allowing transfer through instance selection and reweighting. Here, by contrast, the original labeled source data are not assumed to be available; instead, transfer must proceed from an existing source fragility formulation whose functional form and parameter values are known. This setting is common in fragility practice, where published or previously developed models are available for reuse, but the underlying simulation or empirical datasets used to build them are not accessible. In such cases, adaptation cannot rely on source-instance reweighting and must instead proceed by recalibrating the source model parameters using the limited observations available in the target region.

A further challenge in this case study is that the target observations are sparse and heterogeneous across meaningful subgroups. In practical terms, this occurs when post-event observations are distributed unevenly across categories such as structural classes or other strata. Under these conditions, direct subgroup-specific recalibration may be unstable, particularly when some groups contain only a small number of damage observations. This motivates a complementary hierarchical Bayesian transfer learning formulation, which allows subgroup-specific parameters to borrow strength from one another while preserving uncertainty quantification.

\paragraph{Fine-tuning formulation.}

Let $\boldsymbol{\theta}_{\mathrm{src}}^{*}$ denote the fitted coefficient vector of the source fragility model, and let $\mathcal{D}_T$ denote the labeled target dataset. Here, $\boldsymbol{\theta}$ refers to the model coefficients (e.g., intercept and regression weights in the fragility function). The fine-tuned coefficient vector is expressed as $\boldsymbol{\theta}=(\boldsymbol{\theta}_{\mathrm{b}},\boldsymbol{\theta}_{\mathrm{e}})$, where $\boldsymbol{\theta}_{\mathrm{b}}$ contains the updated coefficients associated with the original base-model terms and $\boldsymbol{\theta}_{\mathrm{e}}$ contains optional target-specific extension coefficients associated with additional local covariates not included in the original fragility function. Model adaptation is performed by minimizing the regularized objective

\begin{equation}
\mathcal{L}(\boldsymbol{\theta})=
\mathcal{L}_{\mathrm{w}}(\boldsymbol{\theta};\mathcal{D}_T)
+\frac{\mu_{\mathrm{b}}}{2}\left\lVert \boldsymbol{\theta}_{\mathrm{b}}-\boldsymbol{\theta}_{\mathrm{src}}^{*}\right\rVert_2^2
+\frac{\mu_{\mathrm{e}}}{2}\left\lVert \boldsymbol{\theta}_{\mathrm{e}}\right\rVert_2^2,
\label{eq:case2_objective}
\end{equation}

\noindent where $\mathcal{L}_{\mathrm{w}}$ is a class-weighted binary cross-entropy loss, $\mu_{\mathrm{b}}$ controls the penalty on deviations from the source-model parameters, and $\mu_{\mathrm{e}}$ controls shrinkage of target-specific extension parameters toward zero. This formulation enables the fine-tuned model to benefit from source-domain knowledge while limiting overfitting under sparse target observations.

\subsubsection{Illustrative example: Hurricane fragility adaptation for residential buildings}
\label{sec:case2_example}

\paragraph{Base model.}

The base model considered herein is the multi-hazard residential-building fragility formulation developed by \textcite{do_hurricane_2020} for wood-framed, single-story buildings under combined storm surge and wave loading. The failure probability is expressed as

\begin{equation}
P_f(d,w)=\frac{Z(d,w)^{\beta}}{Z(d,w)^{\beta}+1},
\label{eq:case2_fragility}
\end{equation}

where the combined hazard index $Z(d,w)$ is defined as

\begin{equation}
Z(d,w)=\frac{d}{A}+\frac{w}{B}+C\left(\frac{d}{A}\right)\left(\frac{w}{B}\right).
\label{eq:case2_hazard_index}
\end{equation}

In Equations~\eqref{eq:case2_fragility} and \eqref{eq:case2_hazard_index}, $d$ denotes storm-surge depth, $w$ denotes wave height, $A$ and $B$ are scaling parameters controlling the influence of surge and wave intensity respectively, $C$ is an interaction parameter capturing the coupled surge--wave effect, and $\beta$ is a shape parameter controlling the steepness of the fragility response. The parameter set $(A,B,C,\beta)$ varies across First-Floor Elevation (FFE) categories in the base model, and this subgroup-specific structure is preserved in the adaptation.

\paragraph{Target region, data, and challenges.}

The target domain comprises Hurricane Ian (2022) residential-building damage observations from Fort Myers, Florida \parencite{figueira2025virtual}, paired with ADCIRC/SWAN numerical model hindcast estimates of surge depth and wave height \parencite{kaiser2023adcirc}. To maintain consistency with source model assumptions, the analysis is restricted to single-story wood-framed residences. The target dataset is stratified by First-Floor Elevation (FFE) category, with the response variable defined as complete structural failure. The target data present several simultaneous challenges: stratification by FFE category with unequal subgroup sizes, strong class imbalance with differing failure proportions across FFE bins, and limited sample sizes within several elevation categories. FFE categories with fewer than 10 observations are excluded from the analysis to reduce estimation instability.

\paragraph{Hyperparameter tuning and cross-validation.}

Because class balance and sample size vary substantially across FFE categories, hyperparameter tuning accounts for subgroup heterogeneity. Stratified $K$-fold cross-validation is employed for categories with sufficient and reasonably balanced samples, whereas leave-one-out cross-validation is used for sparse categories to maximize use of limited observations. To prevent data-rich FFE groups from dominating the results, overall performance is aggregated using sample-weighted metrics across categories. This stratified procedure is essential because the fragility model parameters are FFE-dependent, and transfer learning benefits vary across elevation bins.

\paragraph{Extended covariates and feature selection.}

Beyond the two hazard variables (surge depth and wave height) used in the source model, the target dataset includes building-specific attributes (building age, distance to shoreline, and footprint area) that may capture local vulnerability characteristics not represented in the simulation-derived source model. Rather than including all variables by default, a structured feature-selection procedure is performed prior to model fine-tuning: candidate feature subsets are evaluated within each FFE category using cross-validation, and performance is aggregated across categories using sample-weighted F1 scores. The feature subset achieving the highest weighted F1 is selected for the extended model. When local covariates are included, the hazard index is augmented as

\begin{equation}
Z_{\mathrm{ext}}(d,w,x_1,x_2,x_3)
=
Z(d,w)+\gamma_{1}x_{1}+\gamma_{2}x_{2}+\gamma_{3}x_{3},
\label{eq:case2_extended_index}
\end{equation}

where $x_1$, $x_2$, and $x_3$ denote standardized building age, distance to shoreline, and footprint area, respectively, and $\gamma_1$, $\gamma_2$, $\gamma_3$ are the corresponding extension coefficients. Depending on feature selection results, only a subset of these terms is retained in the final model.

\paragraph{Hierarchical Bayesian formulation.}

Although deterministic fine-tuning substantially improves predictive performance, some FFE categories remain too sparse for stable independent calibration. Following the approach of \textcite{saeednejad2025bridging}, hierarchical Bayesian transfer learning addresses this limitation by enabling partial pooling across FFE categories and providing rigorous uncertainty quantification. The central principle is to treat subgroup-specific fragility parameters as related but not identical, allowing data-rich groups to inform estimation in data-poor groups through shared hyperdistributions.

Let $g_i\in\{1,\dots,G\}$ denote the FFE category of observation $i$. For each category, the parameter vector $\boldsymbol{\theta}_g=\{A_g,B_g,C_g,\beta_g\}$ is modeled hierarchically, with group-level parameters drawn from common hyperdistributions whose means are informed by the source fragility model. The failure probability for observation $i$ is expressed as

\begin{equation}
P_{f,i}=\sigma\!\left(\beta_{g_i}\ln Z_i\right),
\label{eq:case2_bayes_failure_prob}
\end{equation}

\noindent where $\sigma(\cdot)$ denotes the standard logistic (sigmoid) function $\sigma(t)=1/(1+e^{-t})$, and

\begin{equation}
Z_i=
\frac{d_i}{A_{g_i}}
+
\frac{w_i}{B_{g_i}}
+
C_{g_i}
\left(\frac{d_i}{A_{g_i}}\right)
\left(\frac{w_i}{B_{g_i}}\right).
\label{eq:case2_bayes_hazard_index}
\end{equation}

In Equations~\eqref{eq:case2_bayes_failure_prob} and \eqref{eq:case2_bayes_hazard_index}, $d_i$ and $w_i$ denote surge depth and wave height for observation $i$, while $A_{g_i}$, $B_{g_i}$, $C_{g_i}$, and $\beta_{g_i}$ are the FFE-specific fragility parameters for the subgroup to which observation $i$ belongs. To preserve physical interpretability, all parameters are constrained to remain positive through appropriate prior specifications (e.g., log-normal priors). Hyperpriors on subgroup means are centered on source-model values, thereby encoding transferable knowledge into the Bayesian hierarchy.

\paragraph{Weighted likelihood and posterior inference.}

To address class imbalance within the Bayesian framework, the Bernoulli likelihood is weighted so that failure observations contribute more strongly to posterior updating. Let $y_i\in\{0,1\}$ denote the observed binary damage outcome for observation $i$, and let $P_{f,i}$ denote the corresponding modeled failure probability. The likelihood for each observation is

\begin{equation}
y_i \sim \mathrm{Bernoulli}(P_{f,i}),
\label{eq:case2_bernoulli}
\end{equation}

and the weighted log-likelihood is

\begin{equation}
\ln \mathcal{L}_{\mathrm{w}}
=
\sum_i \omega_i \bigl[y_i \ln P_{f,i} + (1-y_i)\ln(1-P_{f,i})\bigr],
\label{eq:case2_weighted_likelihood}
\end{equation}

\noindent where $\omega_i$ in Equation~\eqref{eq:case2_weighted_likelihood} denotes the observation weight for sample $i$, derived from inverse class frequency to up-weight the minority (failure) class; the per-observation Bernoulli likelihood of Equation~\eqref{eq:case2_bernoulli} is reweighted accordingly. Posterior inference is performed using Markov Chain Monte Carlo sampling with the No-U-Turn Sampler (NUTS) \parencite{hoffman2014nuts}, yielding posterior distributions for FFE-specific parameter sets and posterior predictive uncertainty. This formulation is particularly valuable for sparse categories where deterministic point estimates may be unstable.

\paragraph{Results and discussion.}

\begin{figure}[pos=htbp]
  \centering
  \begin{subfigure}[t]{0.49\textwidth}
    \centering
    \includegraphics[width=\textwidth]{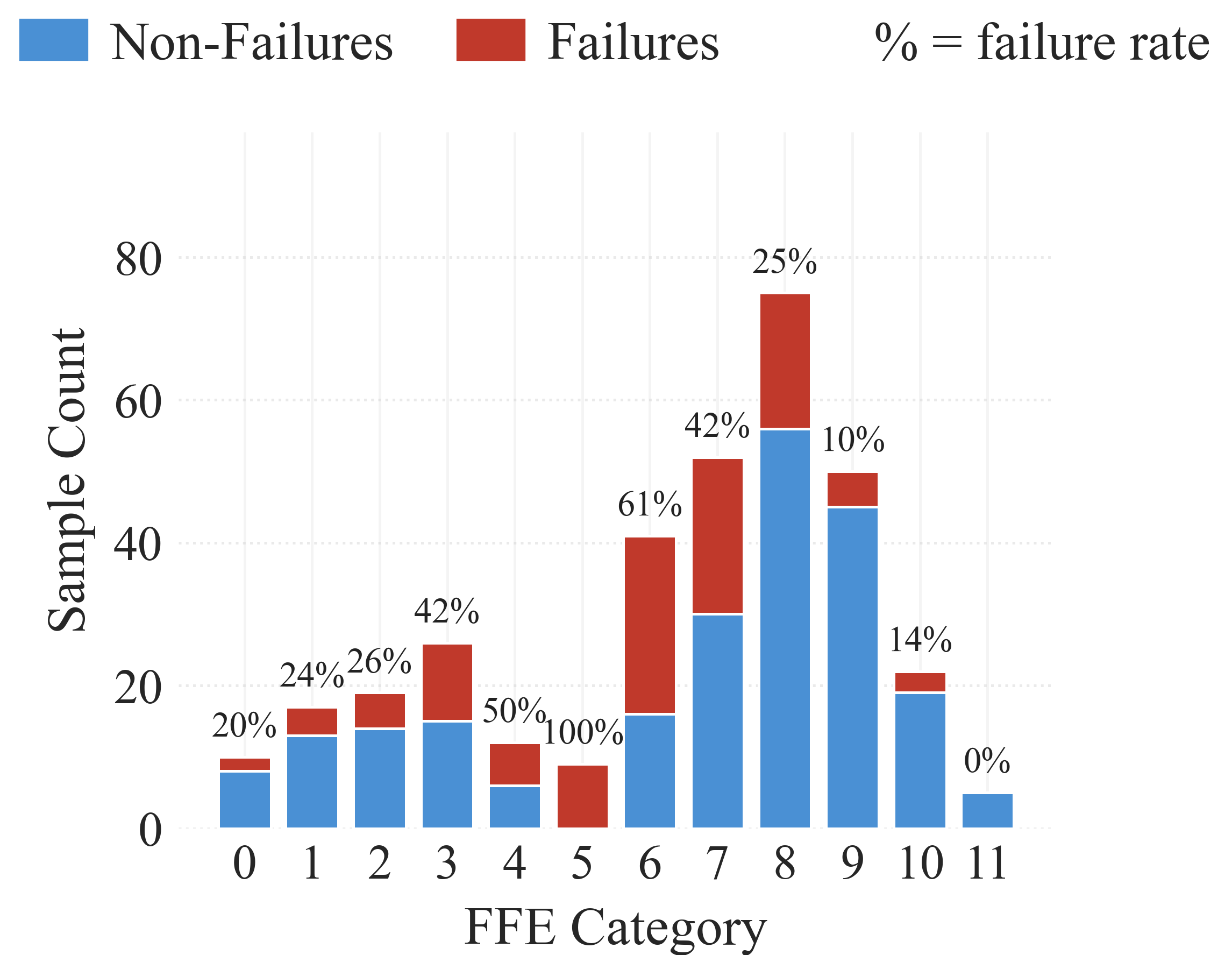}
    \caption{Failures and non-failures by FFE category.}
    \label{fig:case2_imbalance}
  \end{subfigure}\hfill
  \begin{subfigure}[t]{0.49\textwidth}
    \centering
    \includegraphics[width=\textwidth]{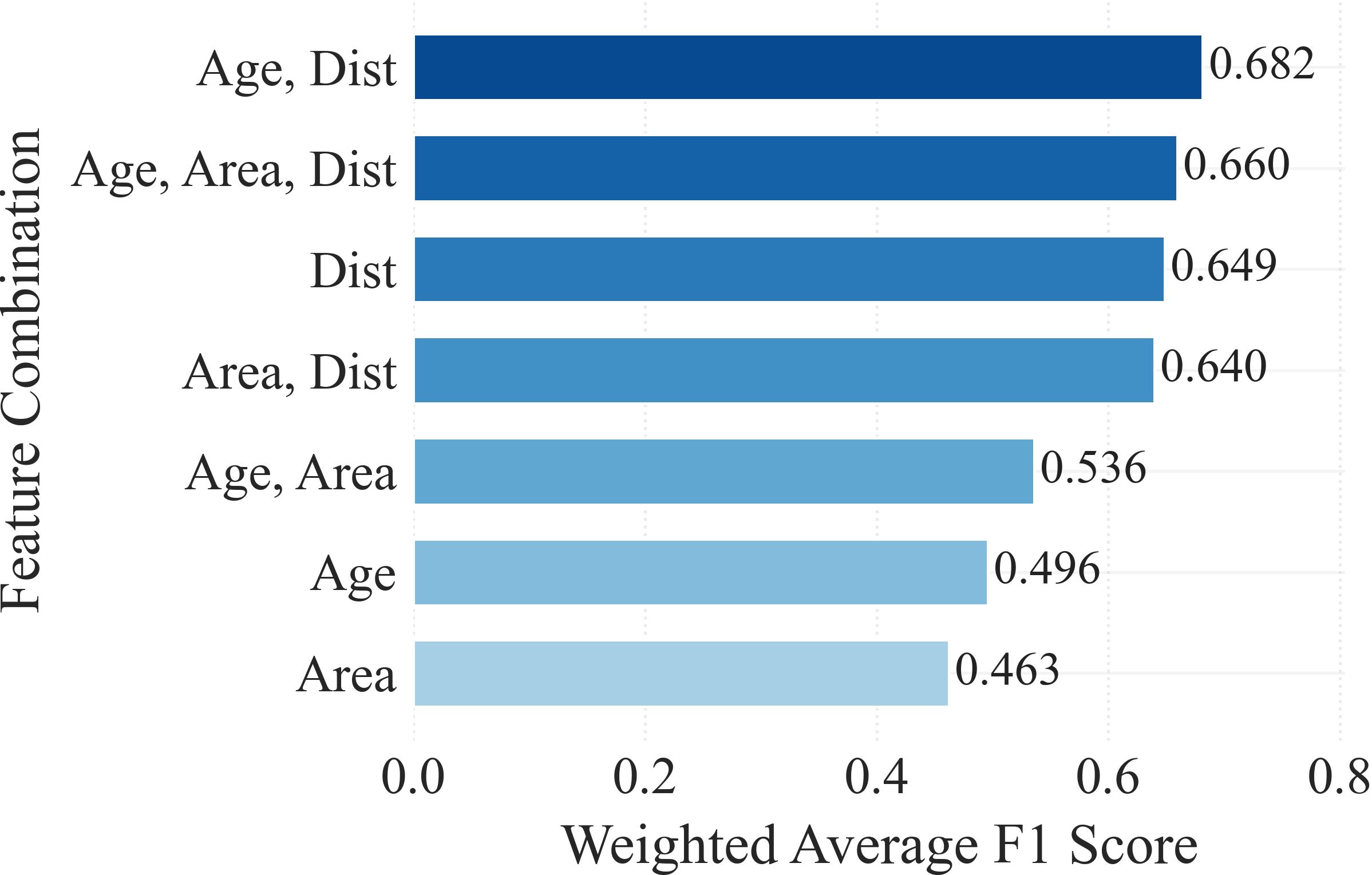}
    \caption{Weighted F1 scores for candidate feature subsets.}
    \label{fig:case2_featselect}
  \end{subfigure}
  \caption{Case Study II: target-data imbalance and extended-feature selection diagnostics.}
\end{figure}

The FFE-specific class imbalance in the target dataset is illustrated in Figure~\ref{fig:case2_imbalance}, highlighting the uneven distribution of failures and non-failures across elevation bins. Figure~\ref{fig:case2_featselect} summarizes the weighted-F1 comparison of candidate extended-feature subsets. In particular, Figure~\ref{fig:case2_featselect} shows how predictive performance changes as different combinations of candidate local covariates are added to the base model during feature-selection testing. Higher weighted-F1 values indicate a better balance between correctly identifying both failure and non-failure classes while accounting for class imbalance. The superior performance of the subset including building age and distance to shoreline indicates that these variables provide the most useful complementary local information among the candidates considered, motivating their inclusion in the extended fine-tuned model. Fine-tuning is performed by minimizing Equation~\eqref{eq:case2_objective}, with the extended hazard index of Equation~\eqref{eq:case2_extended_index} substituted in the objective when local covariates are included, using bound-constrained optimization to enforce physically meaningful parameter ranges.

\begin{figure}[pos=htbp]
  \centering
  \includegraphics[width=\stdfigurewidth]{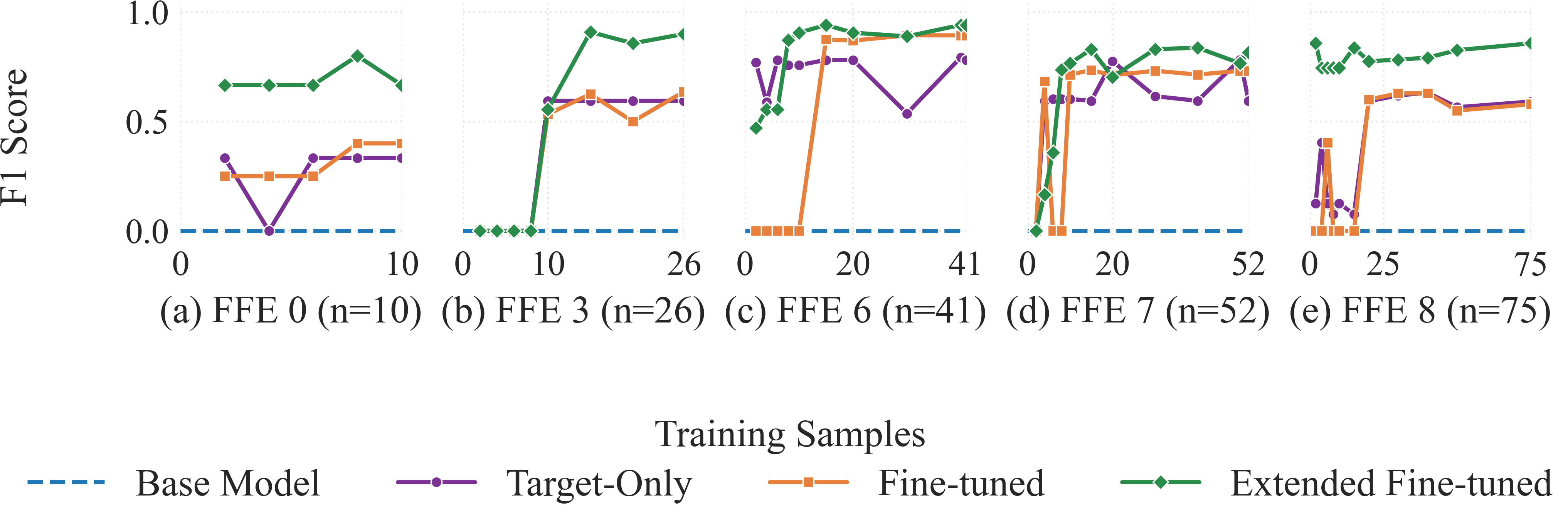}
  \caption{Case Study II: incremental F1 score versus target training sample size for five representative FFE categories (0, 3, 6, 7, 8).}
  \label{fig:case2_incremental_f1}
\end{figure}

To evaluate data efficiency, model performance is examined as a function of target training sample size within each FFE category. Figure~\ref{fig:case2_incremental_f1} shows the trend for five representative FFE categories (0, 3, 6, 7, and 8) spanning the available sample-size range: direct application of the source model yields near-zero F1 scores in every panel, indicating poor transferability without adaptation. A model trained from scratch on target data alone improves gradually as more target observations become available, whereas transfer learning achieves substantially faster gains. Fine-tuning the source parameters improves performance even in low-data settings, and incorporating selected extended covariates yields further improvements in the better-sampled categories. Each panel's x-axis explicitly marks the maximum available target training sample count for that category.

\begin{figure}[pos=htbp]
  \centering
  \includegraphics[width=\stdfigurewidth]{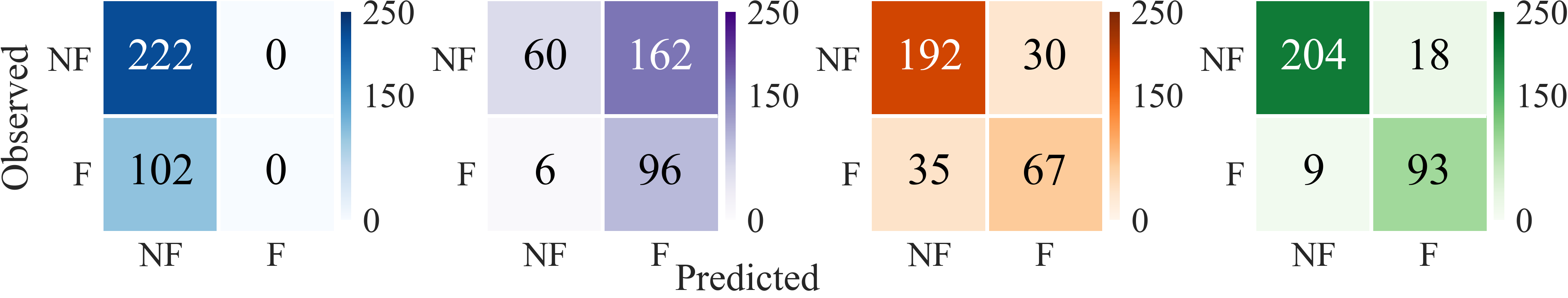}
  \par\vspace{1pt}
  {\small\sffamily\makebox[\stdfigurewidth][c]{%
    \makebox[\dimexpr\stdfigurewidth/4\relax][c]{\textbf{(a)} Base Model}%
    \makebox[\dimexpr\stdfigurewidth/4\relax][c]{\textbf{(b)} Target-Only}%
    \makebox[\dimexpr\stdfigurewidth/4\relax][c]{\textbf{(c)} Fine-tuned}%
    \makebox[\dimexpr\stdfigurewidth/4\relax][c]{\textbf{(d)} Extended Fine-tuned}%
  }}
  \caption{Case Study II: confusion matrices for the four modeling scenarios (NF: not failed; F: failed).}
  \label{fig:case2_confusion_compare}
\end{figure}

A pooled comparison of four modeling scenarios is reported using confusion matrices (Figure~\ref{fig:case2_confusion_compare}) and aggregated classification metrics (Table~\ref{tab:case2_method_compare}): (i) direct application of the base model, (ii) a model trained from scratch using target data only, (iii) transfer learning with fine-tuning of source parameters only, and (iv) transfer learning with both source and extended parameters. The confusion matrices demonstrate that the source model without adaptation fails to identify failures, whereas transfer-learning formulations substantially improve failure detection. Table~\ref{tab:case2_method_compare} confirms this trend quantitatively: the extended fine-tuned model achieves the highest overall accuracy, precision, recall, and F1 score, indicating that selected local building attributes contribute valuable information beyond the original surge--wave hazard representation.

\begin{table}[width=.9\linewidth,cols=5,pos=t]
\caption{Case Study II: pooled classification metrics across FFE categories ($n=324$).}
\label{tab:case2_method_compare}
\begin{tabular*}{\tblwidth}{@{}LCCCC@{}}
\toprule
Method & Accuracy & Precision & Recall & F1 \\
\midrule
Base Model & 0.685 & 0.000 & 0.000 & 0.000 \\
Target-Only & 0.481 & 0.372 & 0.941 & 0.533 \\
Fine-tuned & 0.799 & 0.691 & 0.657 & 0.673 \\
\textbf{Extended Fine-tuned} & \textbf{0.917} & \textbf{0.838} & \textbf{0.912} & \textbf{0.873} \\
\bottomrule
\end{tabular*}
\end{table}

\begin{figure}[pos=htbp]
  \centering
  \includegraphics[width=0.7\linewidth]{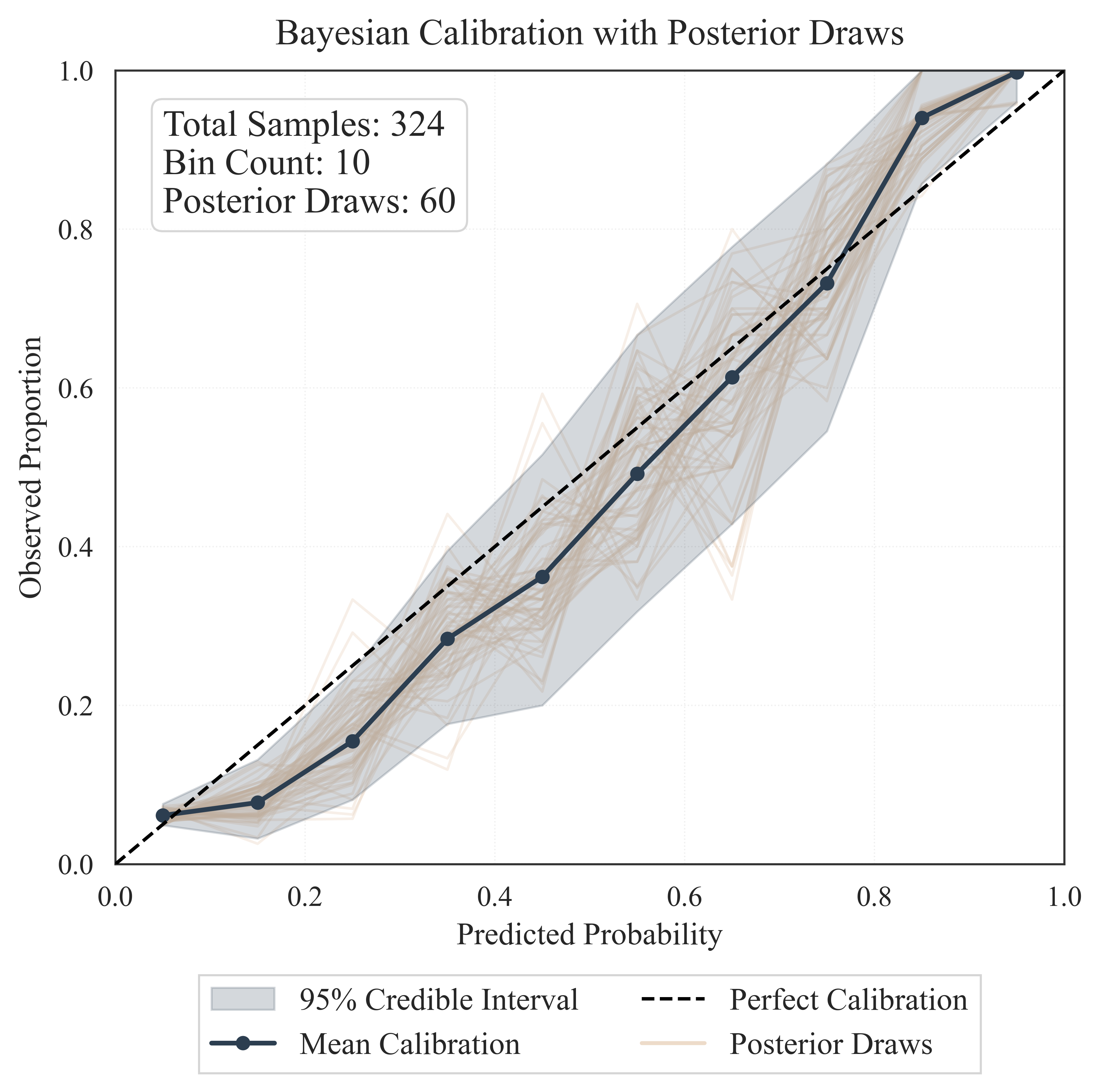}
  \caption{Case Study II: Bayesian calibration curve with posterior draws and 95\% credible band.}
  \label{fig:case2_calibration}
\end{figure}

The hierarchical Bayesian variant stabilizes estimation in sparse FFE categories and quantifies predictive uncertainty. Calibration performance is summarized in Figure~\ref{fig:case2_calibration}, where posterior predictive probabilities are compared against observed failure proportions and the 95\% credible band reflects epistemic uncertainty arising from limited and heterogeneous target data. Relative to deterministic fine-tuning, this Bayesian formulation provides a more comprehensive representation of uncertainty and is therefore particularly valuable when fragility estimates support risk-informed decision-making.

% -------------------------------------------------
\subsection{Case Study III: Multi-source transfer learning for seismic fragility adaptation}
\label{sec:case3}

\subsubsection{Methodology}
\label{sec:case3_methodology}

\paragraph{Problem setting.}

This case study addresses a third transfer setting that is distinct from those considered in Case Studies I and II. Here, multiple candidate source fragility models are available for the same structural class, yet none can be assumed a priori to transfer reliably to the target domain in isolation. This situation commonly arises when fragility libraries have been developed across different studies, regions, or modeling assumptions, yielding several plausible source models without a clear basis for selecting a single best model for the target setting.

Under these conditions, direct adoption of any one source model may result in biased or unstable predictions, while selecting a single source model in advance may forfeit complementary information contained in the remaining models. Multi-source transfer learning is therefore introduced to combine information across candidate source fragility models through learned relative weights, while permitting regularized target-specific adjustments. The resulting formulation enables the target model to leverage complementary source knowledge while limiting over-reliance on any single source and controlling overfitting under limited target observations.

\paragraph{Multi-source fusion formulation.}

Although fragility models in the literature may be expressed in different forms, the proposed fusion framework is not limited to lognormal fragilities. Its main requirement is that source-model information be mapped into a common parameterized representation so that multiple sources can be combined consistently. In this case study, that common representation is taken to be lognormal fragility parameters. Accordingly, assume that $m$ source fragility models provide lognormal fragility parameters $\{\lambda_k^{(i)},\zeta_k^{(i)}\}$ for each damage-state threshold $k$. Source contributions are governed by softmax-constrained weights:
\begin{equation}
w_i=\frac{\exp(\alpha_i)}{\sum_{j=1}^{m}\exp(\alpha_j)}.
\label{eq:case3_softmax_weights}
\end{equation}
To preserve parameter positivity and physical plausibility, fusion is performed on the log scale and augmented by learnable deviation terms:
\begin{align}
\ln(\lambda_{k,\mathrm{TL}}) &= \sum_{i=1}^{m} w_i\,\ln(\lambda_k^{(i)}) + \Delta\lambda_k,
\label{eq:case3_lambda_fusion}\\
\ln(\zeta_{k,\mathrm{TL}}) &= \sum_{i=1}^{m} w_i\,\ln(\zeta_k^{(i)}) + \Delta\zeta_k.
\label{eq:case3_zeta_fusion}
\end{align}
The softmax weights $w_i$ of Equation~\eqref{eq:case3_softmax_weights} provide transparent interpretation of how the framework allocates trust across imperfect sources, while the deviation terms $\Delta\lambda_k$ and $\Delta\zeta_k$ in Equations~\eqref{eq:case3_lambda_fusion} and \eqref{eq:case3_zeta_fusion} capture target-specific corrections beyond the weighted combination of source models.

\paragraph{Regularized optimization.}

Weights and deviation terms are estimated by minimizing a regularized negative log-likelihood:
\begin{equation}
\mathcal{L}_{\mathrm{MS}}
=
-\ln p(\mathcal{D}_T \mid \{\lambda_{k,\mathrm{TL}},\zeta_{k,\mathrm{TL}}\})
+R_1(\Delta\lambda,\Delta\zeta)
+R_2(\Delta\lambda,\Delta\zeta)
+R_3(\mathbf{w}),
\label{eq:case3_objective}
\end{equation}
where, in Equation~\eqref{eq:case3_objective}, $\mathbf{w}=[w_1,\dots,w_m]^\top$ denotes the vector of source-fusion weights, $R_1$ penalizes large deviations to ensure stability, $R_2$ promotes parsimonious corrections across damage states to enhance interpretability, and $R_3$ discourages over-reliance on a single source by regularizing the fusion weights. Regularization strengths are tuned via cross-validation on the target data.

\subsubsection{Illustrative example: Seismic fragility adaptation for reinforced concrete bridges}
\label{sec:case3_example}

\paragraph{Base models.}

Three base fragility models are considered, derived from prior simulation-based seismic fragility studies for reinforced concrete bridge classes \parencite{mangalathu2017hazus,mangalathu2017thesis,chen2025secondgen}.The base fragility models used in this study were developed through related but distinct simulation-based workflows. \textcite{mangalathu2017thesis} develops fragility relationships through a performance-based grouping framework in which bridge subclasses are identified from nonlinear analytical response and then used for fragility estimation. By contrast, \textcite{mangalathu2017hazus} uses ANOVA-based grouping specifically to refine HAZUS bridge classes and then constructs component- and system-level lognormal fragility curves for the grouped classes. Accordingly, both studies rely on nonlinear analysis and statistical grouping, but they differ in methodological emphasis: one is centered on fragility development after subclass identification, while the other is centered on HAZUS class refinement and explicit system-fragility construction from component responses. This heterogeneity motivates the use of multi-source transfer learning to integrate multiple prior fragility sources while retaining flexibility for target-domain adjustment. Each source model provides lognormal fragility parameters for Era~1 (pre-1970) bridges. The fragility functions follow the standard form
\begin{equation}
P(DS\ge k\mid \mathrm{IM})=\Phi\left(\frac{\ln(\mathrm{IM})-\lambda_k}{\zeta_k}\right),
\label{eq:case3_source_fragility}
\end{equation}
where, in Equation~\eqref{eq:case3_source_fragility}, $\Phi(\cdot)$ is the standard normal cumulative distribution function, $\lambda_k$ and $\zeta_k$ are the lognormal median and dispersion parameters for damage-state threshold $k$, and $\mathrm{IM}=S_{a}(T=1.0\,\mathrm{s})$ is the spectral acceleration at 1.0-second period, selected for compatibility across source models and the target inventory. Table~\ref{tab:case3_source_params} lists the base fragility parameters used in multi-source fusion.

\paragraph{Target region, data, and challenges.}

The target domain comprises empirical damage observations from the 2001 Nisqually earthquake for Era~1 (pre-1970) reinforced concrete bridges in Washington State \parencite{ranf_damage_nodate}. These observations represent real post-earthquake inspections with documented damage states, providing valuable empirical evidence for validating and adapting analytical fragility models. Several challenges are present: the three analytical source models differ in their calibration assumptions and regional biases, none is expected to transfer reliably in isolation, the target region offers only limited empirical observations, and damage state classifications from post-earthquake inspections carry inherent uncertainty and potential misclassification.

\begin{table}[width=\linewidth,cols=7,pos=htbp]
\caption{Case Study III: base and adapted fragility parameters for Era~1 (pre-1970) bridges, $S_{a}(T=1.0\,\mathrm{s})$.}
\label{tab:case3_source_params}
\small
\renewcommand{\arraystretch}{1.2}
\begin{tabular*}{\tblwidth}{@{}LCCCCCC@{}}
\toprule
& \multicolumn{2}{c}{Slight} 
& \multicolumn{2}{c}{Moderate} 
& \multicolumn{2}{c}{Complete} \\
\cmidrule(lr){2-3}\cmidrule(lr){4-5}\cmidrule(lr){6-7}
Source model 
& $\lambda$ & $\zeta$ 
& $\lambda$ & $\zeta$ 
& $\lambda$ & $\zeta$ \\
\midrule
Base Model 1 \parencite{mangalathu2017thesis} & 0.05 & 0.62 & 0.11 & 0.62 & 0.31 & 0.59 \\
Base Model 2 \parencite{mangalathu2017hazus} & 0.09 & 0.52 & 0.20 & 0.53 & 0.58 & 0.53 \\
Base Model 3 \parencite{mangalathu2017hazus} & 0.09 & 0.51 & 0.19 & 0.52 & 0.56 & 0.52 \\
Adapted & 0.02 & 0.54 & 0.27 & 0.57 & 1.17 & 0.54 \\
\bottomrule
\end{tabular*}
\end{table}

\paragraph{Results and discussion.}

Learned source weights presented in Table~\ref{tab:case3_source_weights} provide transparent evidence of how the method allocates trust across imperfect analytical models while retaining a stable, physically meaningful baseline. In data-scarce regimes, this interpretability is valuable because it reveals whether adaptation relies primarily on a single source or benefits from balanced fusion across multiple sources.

\begin{table}[width=.5\linewidth,cols=2,pos=t]
\caption{Case Study III: learned multi-source fusion weights for Era~1 bridges.}
\label{tab:case3_source_weights}
\begin{tabular*}{\tblwidth}{@{}LC@{}}
\toprule
Source model & Weight \\
\midrule
Base Model 1 & 0.293 \\
Base Model 2 & 0.322 \\
Base Model 3 & 0.385 \\
\bottomrule
\end{tabular*}
\end{table}

\begin{figure}[pos=htbp]
  \centering
  \includegraphics[width=\stdfigurewidth]{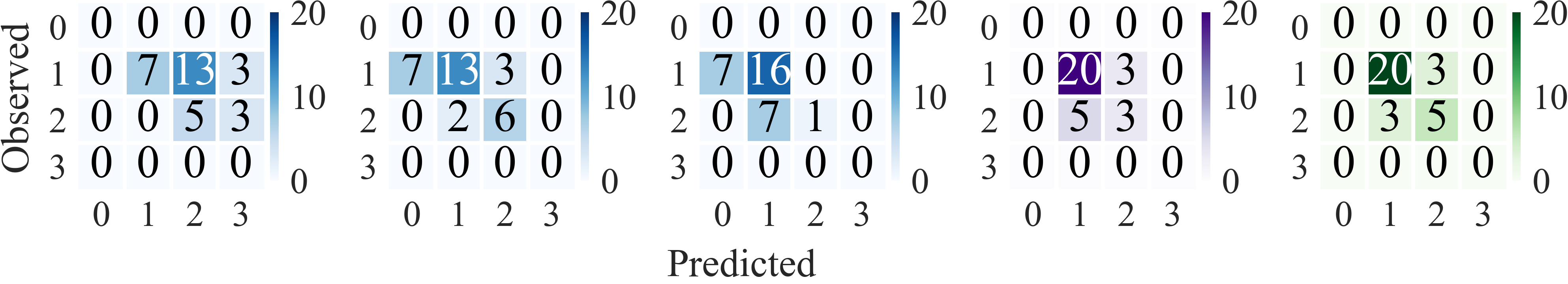}
  \par\vspace{1pt}
  {\small\sffamily\makebox[\stdfigurewidth][c]{%
    \makebox[\dimexpr\stdfigurewidth/5\relax][c]{\textbf{(a)} Base Model 1}%
    \makebox[\dimexpr\stdfigurewidth/5\relax][c]{\textbf{(b)} Base Model 2}%
    \makebox[\dimexpr\stdfigurewidth/5\relax][c]{\textbf{(c)} Base Model 3}%
    \makebox[\dimexpr\stdfigurewidth/5\relax][c]{\textbf{(d)} Target-Only}%
    \makebox[\dimexpr\stdfigurewidth/5\relax][c]{\textbf{(e)} Adapted}%
  }}
  \caption{Case Study III: confusion matrices for Era~1 (Pre-1970) bridges across the five models (damage-state classes $0$--$3$).}
  \label{fig:case3_confusion_s3}
\end{figure}

Performance comparison across base models, a target-only baseline trained without transfer learning, and the adapted multi-source model is presented in Figure~\ref{fig:case3_confusion_s3} and Table~\ref{tab:case3_confusion_metrics}. The adapted model achieves substantially improved macro-averaged precision, recall, and F1 scores relative to both the strongest individual source model and the target-only baseline, demonstrating the benefit of multi-source fusion with regularized target-specific corrections under limited empirical observations.

\begin{table}[width=.9\linewidth,cols=5,pos=t]
\caption{Case Study III: macro-averaged classification metrics for Era~1 bridges.}
\label{tab:case3_confusion_metrics}
\begin{tabular*}{\tblwidth}{@{}LCCCC@{}}
\toprule
Method & Accuracy & Precision & Recall & F1 \\
\midrule
Base Model 1 & 0.387 & 0.426 & 0.310 & 0.284 \\
Base Model 2 & 0.613 & 0.511 & 0.438 & 0.463 \\
Base Model 3 & 0.548 & 0.565 & 0.274 & 0.306 \\
Target-Only & 0.742 & 0.650 & 0.622 & 0.631 \\
\textbf{Adapted} & \textbf{0.806} & \textbf{0.747} & \textbf{0.747} & \textbf{0.747} \\
\bottomrule
\end{tabular*}
\end{table}

\begin{figure}[pos=htbp]
  \centering
  \includegraphics[width=\stdfigurewidth]{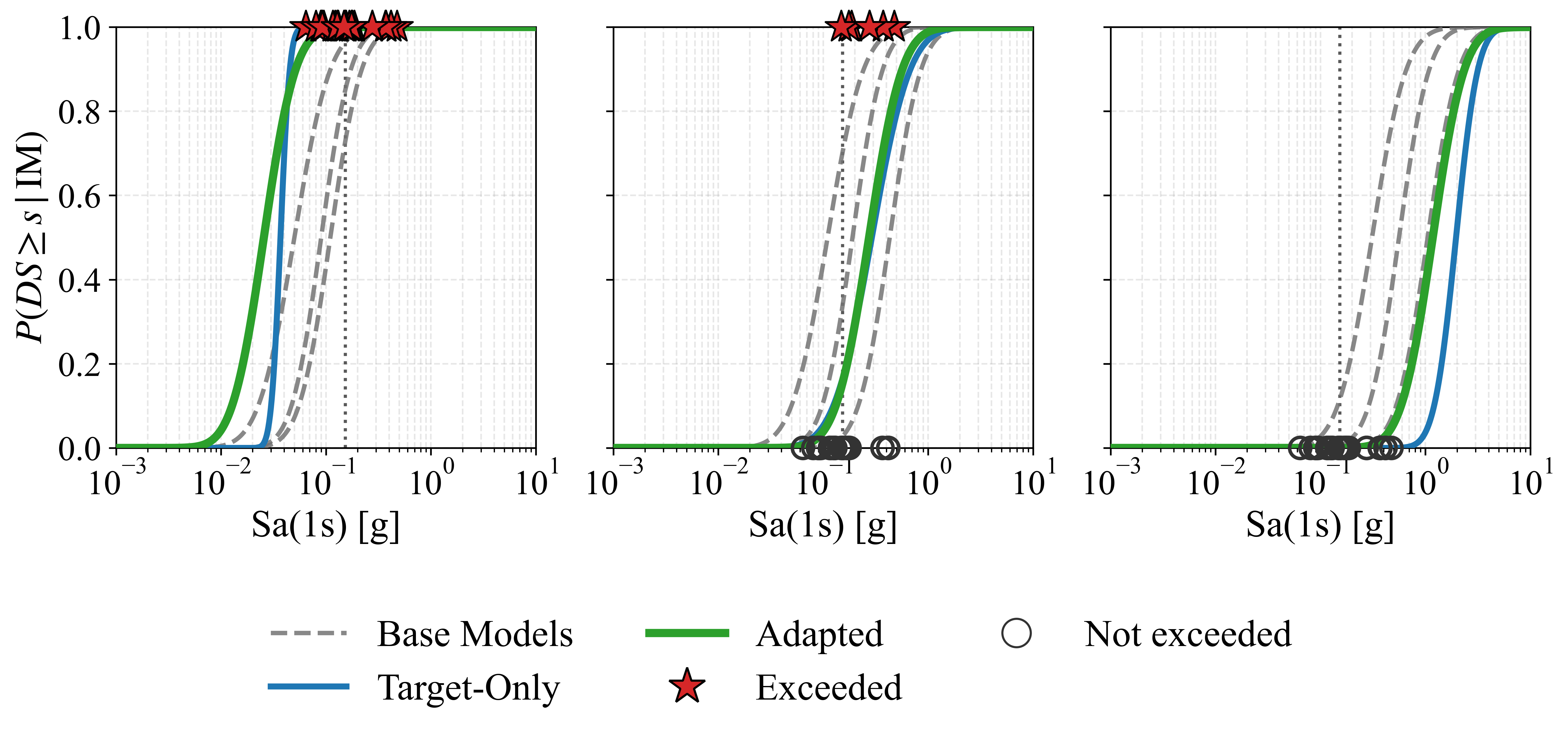}
  \par\vspace{1pt}
  {\small\sffamily\makebox[\stdfigurewidth][c]{%
    \makebox[\dimexpr\stdfigurewidth/3\relax][c]{\textbf{(a)} Slight}%
    \makebox[\dimexpr\stdfigurewidth/3\relax][c]{\textbf{(b)} Moderate}%
    \makebox[\dimexpr\stdfigurewidth/3\relax][c]{\textbf{(c)} Complete}%
  }}
  \caption{Case Study III: marginal fragility curves for Era~1 (Pre-1970) bridges.}
  \label{fig:case3_fragility_overlay_s3}
\end{figure}

The fragility-curve overlay in Figure~\ref{fig:case3_fragility_overlay_s3} clarifies the mechanism behind the classification-metric gains reported above. The three same-era source curves disagree substantially, particularly at lower intensities for the Slight state, reflecting the variability among existing analytical fragility models for Pre-1970 inventory. The target-only baseline tracks the empirical exceedances more closely than any individual source, confirming that the Era~1 observations do contain identifiable failure-rate information, but its curves remain visibly unsmoothed and shift abruptly between damage states owing to the small sample size. The multi-source adapted model, by contrast, retains the physically motivated shape inherited from the source library while adjusting median intensities to align with the observed exceedances, producing curves that interpolate cleanly across damage states and intensity levels. This visual evidence supports the interpretation that the regularized softmax fusion provides a principled middle ground between rigid source reuse and overfitting to scarce target data.

\begin{figure}[pos=htbp]
  \centering
  \includegraphics[width=\stdfigurewidth]{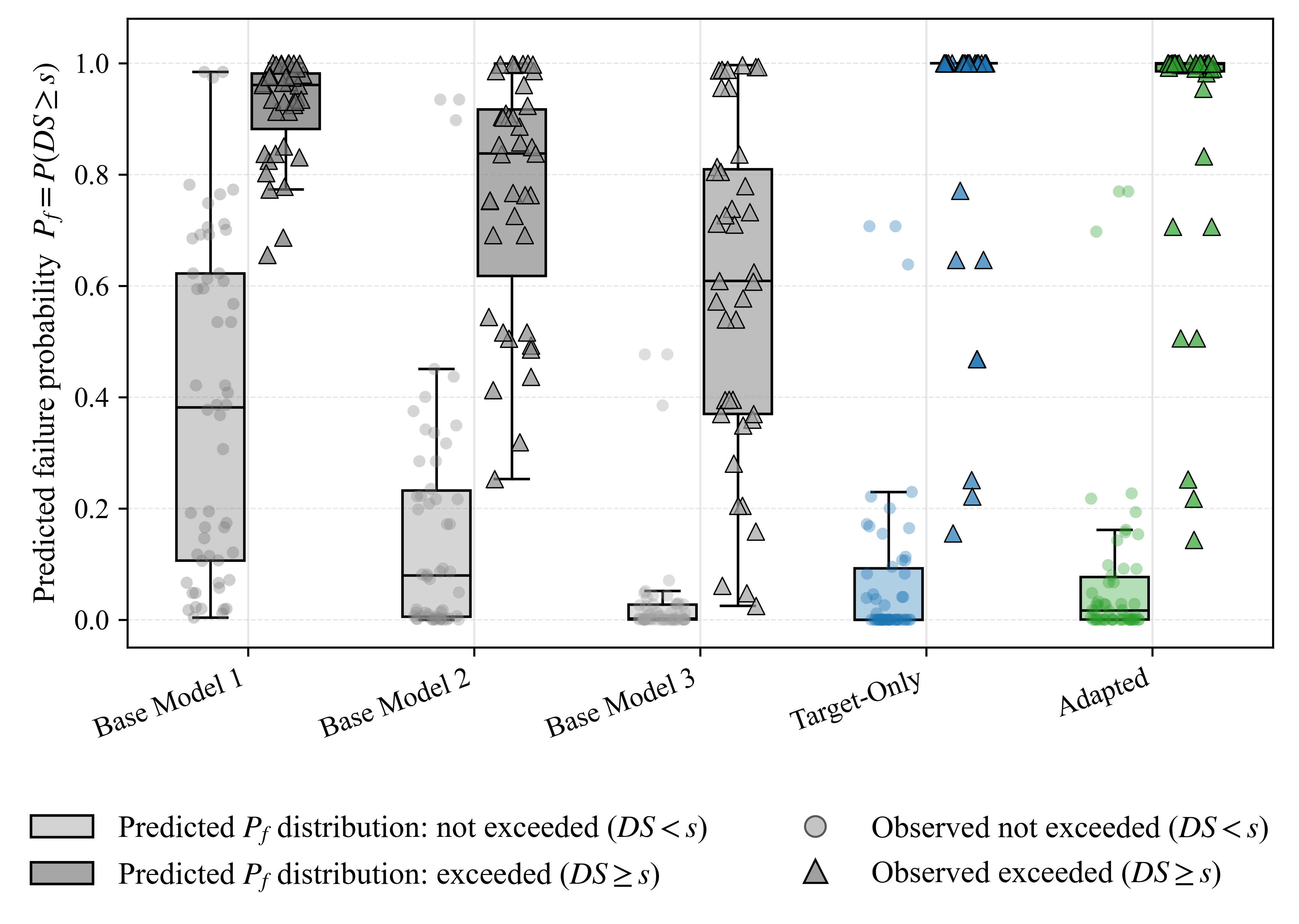}
  \caption{Case Study III: predicted failure probability $P_f=P(DS\ge s)$ for Era~1 (Pre-1970) bridges, pooled over damage states and split by observed outcome (light box: not exceeded; dark box: exceeded).}
  \label{fig:case3_pf_boxplots}
\end{figure}

Figure~\ref{fig:case3_pf_boxplots} summarizes how sharply each model separates exceeded from not-exceeded observations across all damage states. For the three base models the two distributions overlap heavily; Base Models~1 and~2 in particular assign large failure probabilities to many non-exceeded cases, confirming that no single source transfers reliably to the Era~1 target. The Target-Only and Adapted models show the cleanest separation, collapsing the not-exceeded group toward $P_f\approx0$ and the exceeded group toward $P_f\approx1$, with only a handful of misranked high-intensity points. The Adapted model achieves this separation while preserving the smooth, physically grounded fragility shape inherited from the source library, consistent with its higher classification metrics in Table~\ref{tab:case3_confusion_metrics}.

% ========================================
% SECTION 4: DISCUSSION
% ========================================

\section{Discussion: Strategy selection for fragility adaptation}
\label{sec:discussion}

Across the three case studies, fragility adaptation challenges rarely stem from a single limitation. Failures of direct transfer typically arise from coupled mechanisms: limited labeled target observations with sparse failure cases, domain shift between source and target feature distributions, heterogeneity across meaningful subgroups such as FFE categories or design eras, and variability among existing source fragility models. The appropriate adaptation strategy therefore depends jointly on the source and target data regime, the availability and credibility of source functional forms, and the severity of distributional mismatch.

A direct manifestation of these coupled limitations is the failure of unadapted source models in the target domain. Even a physically grounded, simulation-derived fragility model can yield zero predicted failures when deployed without adaptation in a target region exhibiting distributional mismatch and extreme class imbalance (Case Study~II, Table~\ref{tab:case2_method_compare}), and the analogous source surface for the coastal-bridge inventory of Case Study~I produces near-chance discriminative power (AUC $\approx 0.51$, Table~\ref{tab:case1_metrics}). In such settings the headline accuracy may appear acceptable while the model provides no practical value for failure detection. Straightforward pre-transfer diagnostics, such as marginal distribution comparisons and correlation-structure analysis, offer actionable screening tools for identifying when direct transfer is likely unreliable; combined with distribution-level metrics such as MMD computed before and after reweighting, they provide quantitative evidence of whether the selected transfer mechanism meaningfully improves source--target alignment.

When target labels are limited, instance-based and parameter-based adaptation offer two complementary pathways. Instance-based adaptation is most effective when a labeled source dataset exists, target labels are scarce but target features are available, and covariate shift represents the primary barrier to transfer; by reweighting or selecting source instances that are most similar to the target feature distribution, this approach focuses learning on target-relevant regions of the source domain and mitigates negative transfer. Parameter-based adaptation, by contrast, is most appropriate when the functional form of the source fragility model is considered structurally credible but its coefficients require local adjustment due to changes in inventory characteristics or hazard representation. Case Study~II shows that regularized fine-tuning produces substantial gains under limited target data when deviations from the source model are explicitly constrained, and that systematic extended-feature selection further improves performance while preserving interpretability when additional target-region descriptors are available, provided evaluation relies on failure-sensitive metrics and stratified validation protocols that account for subgroup heterogeneity.

Hierarchical Bayesian transfer learning extends parameter-based adaptation when target observations are not only scarce but also heterogeneous across subgroups. Partial pooling lets data-rich strata inform estimation in data-limited strata (in Case Study~II, across FFE categories), while posterior predictive distributions and credible intervals deliver explicit uncertainty quantification suitable for calibration-oriented diagnostics. These benefits come at the cost of additional modelling choices, including prior and hyperprior specification, and increased computational effort, but Case Study~II indicates that the added complexity is warranted when subgroup-level inference is required and when the decision context demands calibrated probabilities rather than deterministic outputs.

Multi-source transfer learning addresses a different regime, in which several candidate source fragility models exist but none is expected to transfer reliably to the target domain in isolation. Case Study~III illustrates that softmax-constrained fusion yields an interpretable allocation of trust across candidate sources, while regularized deviation terms enable empirical correction without overfitting to limited target observations. This scenario is increasingly common as fragility libraries expand across regions, hazards, and structural design eras, making scalable multi-source updating a practical pathway for maintaining library consistency while enabling data-driven refinement whenever empirical evidence becomes available.

% ========================================
% SECTION 5: CONCLUSIONS
% ========================================

\section{Conclusions}
\label{sec:conclusions}

This paper presents a unified transfer learning framework for bridging persistent data gaps in fragility modelling. The framework covers four complementary strategies (instance-based, parameter-based, hierarchical Bayesian, and multi-source) and pairs them, through an explicit decision tree, with the available source knowledge, the size and composition of the target observations, and the severity of source--target mismatch. The strategies are demonstrated through three case studies spanning coastal bridges and residential buildings under hurricane loading and reinforced concrete bridges under seismic loading.

Across all three applications, direct deployment of source fragility models without adaptation proves unreliable when domain shift and severe class imbalance are present, while appropriately selected transfer strategies substantially improve failure detection and predictive stability in low-data regimes. On the Hurricane Katrina coastal-bridge inventory, instance-based reweighting and selection lift the cross-validated AUC from $0.51$ (direct transfer of the source model) to $0.72$ (Table~\ref{tab:case1_metrics}) and recover physically meaningful coefficients from a target dataset with only five observed failures. On the Hurricane Ian residential-building inventory, parameter-based fine-tuning of the source coefficients combined with extended local covariates raises the pooled F1 score from $0.00$ (direct transfer) and $0.53$ (target-only) to $0.87$ (Table~\ref{tab:case2_method_compare}), and the hierarchical Bayesian variant additionally returns calibrated uncertainty bounds in sparse FFE categories. On the Pre-1970 Nisqually seismic-bridge inventory, multi-source fusion of three analytical fragility models with regularised target-specific correction lifts the macro-averaged F1 from $0.46$ for the strongest individual source and $0.63$ for the target-only baseline to $0.75$ (Table~\ref{tab:case3_confusion_metrics}). These results support a systematic workflow that couples domain-shift diagnostics, imbalance-aware training, and interpretable transfer mechanisms tailored to each adaptation challenge.

Several directions remain for future work. Extending the proposed workflow to multi-target settings, where adaptation is performed simultaneously across multiple regions, would improve scalability for regional and portfolio-level risk assessments. Validation across additional hazards and infrastructure classes would broaden the framework's applicability and help fill gaps in existing fragility libraries used for risk and resilience-informed decision-making. Integration with reconnaissance pipelines and remote-sensing damage indicators could further enable closed-loop fragility updating as new post-event evidence accumulates.

\section*{Acknowledgments}

This material is based upon work supported by the National Science Foundation under Award Nos.~2227467 and~2429680.

% ========================================
% CREDIT AUTHORSHIP CONTRIBUTION STATEMENT
% ========================================

\printcredits

\section*{Declaration of generative AI and AI-assisted technologies in the manuscript preparation process}

During the preparation of this work the authors used Claude (Anthropic) in order to perform spelling and grammar checks and to help refine the writing. After using this tool, the authors reviewed and edited the content as needed and take full responsibility for the content of the published article.

% ========================================
% BIBLIOGRAPHY
% ========================================

\printbibliography

@article{balomenos2020parameterized,
  author  = {Balomenos, George P. and Kameshwar, S. and Padgett, Jamie E.},
  title   = {Parameterized fragility models for multi-bridge classes subjected to hurricane loads},
  journal = {Engineering Structures},
  year    = {2020},
  volume  = {208},
  pages   = {110213},
  doi     = {10.1016/j.engstruct.2020.110213}
}

@article{gehl2016development,
  author  = {Gehl, P. and D'Ayala, D.},
  title   = {Development of Bayesian Networks for the multi-hazard fragility assessment of bridge systems},
  journal = {Structural Safety},
  year    = {2016},
  volume  = {60},
  pages   = {37--46},
  doi     = {10.1016/j.strusafe.2016.01.006}
}

@article{karbalayghareh2018optimal,
  author  = {Karbalayghareh, Alireza and Qian, Xiaoning and Dougherty, Edward R.},
  title   = {Optimal Bayesian Transfer Learning},
  journal = {IEEE Transactions on Signal Processing},
  year    = {2018},
  volume  = {66},
  number  = {14},
  pages   = {3724--3739},
  doi     = {10.1109/TSP.2018.2839583}
}

@article{kennedy2001bayesian,
  author  = {Kennedy, Marc C. and O'Hagan, Anthony},
  title   = {Bayesian Calibration of Computer Models},
  journal = {Journal of the Royal Statistical Society: Series B (Statistical Methodology)},
  year    = {2001},
  volume  = {63},
  number  = {3},
  pages   = {425--464},
  doi     = {10.1111/1467-9868.00294}
}

@article{li2013bayesian,
  author  = {Li, Jian and Spencer, Billie F., Jr. and Elnashai, Amr S.},
  title   = {Bayesian Updating of Fragility Functions Using Hybrid Simulation},
  journal = {Journal of Structural Engineering},
  year    = {2013},
  volume  = {139},
  number  = {7},
  pages   = {1160--1171},
  doi     = {10.1061/(ASCE)ST.1943-541X.0000685}
}

@book{little2019missing,
  author    = {Little, Roderick J. A. and Rubin, Donald B.},
  title     = {Statistical analysis with missing data},
  edition   = {3},
  publisher = {Wiley},
  address   = {Hoboken, NJ},
  year      = {2019}
}

@article{padgett2008bridge,
  author  = {Padgett, Jamie and DesRoches, Reginald and Nielson, Bryant and Yashinsky, Mark and Kwon, Oh-Sung and Burdette, Nick and Tavera, Ed},
  title   = {Bridge damage and repair costs from {Hurricane} {Katrina}},
  journal = {Journal of Bridge Engineering},
  year    = {2008},
  volume  = {13},
  number  = {1},
  pages   = {6--14},
  doi     = {10.1061/(ASCE)1084-0702(2008)13:1(6)}
}

@inproceedings{shwartz2022pretrain,
  author    = {Shwartz-Ziv, R. and Goldblum, M. and Souri, H. and Kapoor, S. and Zhu, C. and LeCun, Y. and Wilson, A. G.},
  title     = {Pre-train your loss: easy bayesian transfer learning with informative priors},
  booktitle = {Proceedings of the 36th Conference on Neural Information Processing Systems ({NeurIPS} 2022)},
  year      = {2022}
}

@article{sugiyama2007covariate,
  author  = {Sugiyama, Masashi and Krauledat, Matthias and Muller, K.-R.},
  title   = {Covariate shift adaptation by importance weighted cross validation},
  journal = {Journal of Machine Learning Research},
  year    = {2007},
  volume  = {8},
  pages   = {985--1005}
}

@article{yan2024hierarchical,
  author  = {Yan, Yexiang and Xia, Ye and Sun, Limin},
  title   = {Hierarchical and mixed uncertainty quantification for simulation-based structural seismic fragility analysis},
  journal = {Engineering Structures},
  year    = {2024},
  volume  = {316},
  pages   = {118579},
  doi     = {10.1016/j.engstruct.2024.118579}
}

@article{zhang2022conjugate,
  author  = {Zhang, Yijian and Tien, Iris},
  title   = {Conjugate Bayesian updating of analytical fragility functions using dynamic analysis with application to corroded bridges},
  journal = {Computers and Structures},
  year    = {2022},
  volume  = {270},
  pages   = {106832},
  doi     = {10.1016/j.compstruc.2022.106832}
}

@article{ataei2015fragility,
  author  = {Ataei, Navid and Padgett, Jamie E.},
  title   = {Fragility surrogate models for coastal bridges in hurricane prone zones},
  journal = {Engineering Structures},
  year    = {2015},
  volume  = {103},
  pages   = {203--213},
  doi     = {10.1016/j.engstruct.2015.07.002}
}

@article{do_hurricane_2020,
  author  = {Do, Trung Q. and Van De Lindt, John W. and Cox, Daniel T.},
  title   = {Hurricane Surge-Wave Building Fragility Methodology for Use in Damage, Loss, and Resilience Analysis},
  journal = {Journal of Structural Engineering},
  year    = {2020},
  volume  = {146},
  number  = {1},
  pages   = {04019177},
  doi     = {10.1061/(ASCE)ST.1943-541X.0002472}
}

@techreport{crossett2004population,
  author      = {Crossett, Kristen M. and Culliton, Thomas J. and Wiley, Peter C. and Goodspeed, Timothy R.},
  title       = {Population Trends Along the Coastal United States: 1980--2008},
  institution = {U.S. Department of Commerce, National Oceanic and Atmospheric Administration (NOAA), National Ocean Service},
  address     = {Silver Spring, MD},
  year        = {2004},
  series      = {Coastal Trends Report Series}
}

@article{neumann_future_2015,
  author  = {Neumann, Barbara and Vafeidis, Athanasios T. and Zimmermann, Juliane and Nicholls, Robert J.},
  title   = {Future Coastal Population Growth and Exposure to Sea-Level Rise and Coastal Flooding - {A} Global Assessment},
  journal = {PLOS ONE},
  year    = {2015},
  volume  = {10},
  number  = {3},
  pages   = {e0118571},
  doi     = {10.1371/journal.pone.0118571},
  editor  = {Kumar, Lalit}
}

@article{webster_changes_2005,
  author  = {Webster, P. J. and Holland, G. J. and Curry, J. A. and Chang, H.-R.},
  title   = {Changes in Tropical Cyclone Number, Duration, and Intensity in a Warming Environment},
  journal = {Science},
  year    = {2005},
  volume  = {309},
  number  = {5742},
  pages   = {1844--1846},
  doi     = {10.1126/science.1116448}
}

@article{ellingwood_fragility_2004,
  author  = {Ellingwood, Bruce R. and Rosowsky, David V. and Li, Yue and Kim, Jun Hee},
  title   = {Fragility Assessment of Light-Frame Wood Construction Subjected to Wind and Earthquake Hazards},
  journal = {Journal of Structural Engineering},
  year    = {2004},
  volume  = {130},
  number  = {12},
  pages   = {1921--1930},
  doi     = {10.1061/(ASCE)0733-9445(2004)130:12(1921)}
}

@article{lallemant_statistical_2015,
  author  = {Lallemant, David and Kiremidjian, Anne and Burton, Henry},
  title   = {Statistical procedures for developing earthquake damage fragility curves},
  journal = {Earthquake Engineering \& Structural Dynamics},
  year    = {2015},
  volume  = {44},
  number  = {9},
  pages   = {1373--1389},
  doi     = {10.1002/eqe.2522}
}

@article{porter_creating_2007,
  author  = {Porter, Keith and Kennedy, Robert and Bachman, Robert},
  title   = {Creating Fragility Functions for Performance-Based Earthquake Engineering},
  journal = {Earthquake Spectra},
  year    = {2007},
  volume  = {23},
  number  = {2},
  pages   = {471--489},
  doi     = {10.1193/1.2720892}
}

@techreport{ranf_damage_nodate,
  author      = {Ranf, R. Tyler and Eberhard, Marc O. and Berry, Michael P.},
  title       = {Damage to Bridges during the 2001 {Nisqually} Earthquake},
  institution = {Pacific Earthquake Engineering Research Center, University of California, Berkeley},
  address     = {Berkeley, CA},
  year        = {2001},
  number      = {PEER Report 2001/15}
}

@inproceedings{sun2011twostage,
  author    = {Sun, Qian and Chattopadhyay, Rita and Panchanathan, Sethuraman and Ye, Jieping},
  title     = {A Two-Stage Weighting Framework for Multi-Source Domain Adaptation},
  booktitle = {Advances in Neural Information Processing Systems 24},
  pages     = {505--513},
  year      = {2011}
}

@article{zhuang_comprehensive_2021,
  author  = {Zhuang, Fuzhen and Qi, Zhiyuan and Duan, Keyu and Xi, Dongbo and Zhu, Yongchun and Zhu, Hengshu and Xiong, Hui and He, Qing},
  title   = {A Comprehensive Survey on Transfer Learning},
  journal = {Proceedings of the IEEE},
  year    = {2021},
  volume  = {109},
  number  = {1},
  pages   = {43--76},
  doi     = {10.1109/JPROC.2020.3004555}
}

@article{pan_survey_2010,
  author  = {Pan, Sinno Jialin and Yang, Qiang},
  title   = {A Survey on Transfer Learning},
  journal = {IEEE Transactions on Knowledge and Data Engineering},
  year    = {2010},
  volume  = {22},
  number  = {10},
  pages   = {1345--1359},
  doi     = {10.1109/TKDE.2009.191}
}

@article{mangalathu2017hazus,
  author  = {Mangalathu, Sujith and Soleimani, Farahnaz and Jeon, Jong-Su},
  title   = {Bridge classes for regional seismic risk assessment: Improving HAZUS models},
  journal = {Engineering Structures},
  year    = {2017},
  volume  = {148},
  pages   = {755--766},
  doi     = {10.1016/j.engstruct.2017.07.019}
}

@misc{kouw_introduction_2019,
  author = {Kouw, Wouter M. and Loog, Marco},
  title  = {An Introduction to Domain Adaptation and Transfer Learning},
  year   = {2019}
}

@article{gao2018deep,
  author  = {Gao, Yuqing and Mosalam, Khalid M.},
  title   = {Deep Transfer Learning for Image-Based Structural Damage Recognition},
  journal = {Computer-Aided Civil and Infrastructure Engineering},
  year    = {2018},
  volume  = {33},
  number  = {9},
  pages   = {748--768},
  doi     = {10.1111/mice.12363}
}

@article{ierimonti2021transfer,
  author  = {Ierimonti, Laura and Cavalagli, Nicola and Venanzi, Ilaria and Garc{\'\i}a-Mac{\'\i}as, Enrique and Ubertini, Filippo},
  title   = {A Transfer Bayesian Learning Methodology for Structural Health Monitoring of Monumental Structures},
  journal = {Engineering Structures},
  year    = {2021},
  volume  = {247},
  pages   = {113089},
  doi     = {10.1016/j.engstruct.2021.113089}
}

@article{pan2023transfer,
  author  = {Pan, Qiuyue and Bao, Yuequan and Li, Hui},
  title   = {Transfer Learning-Based Data Anomaly Detection for Structural Health Monitoring},
  journal = {Structural Health Monitoring},
  year    = {2023},
  volume  = {22},
  number  = {5},
  pages   = {3077--3091},
  doi     = {10.1177/14759217221142174}
}

@article{bao2023deep,
  author  = {Bao, Nengxin and Zhang, Tong and Huang, Ruizhi and Biswal, Suryakanta and Su, Jingyong and Wang, Ying},
  title   = {A Deep Transfer Learning Network for Structural Condition Identification with Limited Real-World Training Data},
  journal = {Structural Control and Health Monitoring},
  year    = {2023},
  volume  = {2023},
  pages   = {8899806},
  doi     = {10.1155/2023/8899806}
}

@article{calton2022using,
  author  = {Calton, Landon and Wei, Zhangping},
  title   = {Using Artificial Neural Network Models to Assess Hurricane Damage through Transfer Learning},
  journal = {Applied Sciences},
  year    = {2022},
  volume  = {12},
  number  = {3},
  pages   = {1466},
  doi     = {10.3390/app12031466}
}

@article{zentner_fragility_2017,
  author  = {Zentner, Irmela and G{\"u}ndel, Max and Bonfils, Nicolas},
  title   = {Fragility analysis methods: Review of existing approaches and application},
  journal = {Nuclear Engineering and Design},
  year    = {2017},
  volume  = {323},
  pages   = {245--258},
  doi     = {10.1016/j.nucengdes.2016.12.021}
}

@article{mishra2017hurricane,
  author  = {Mishra, Spandan and Vanli, O. Arda and Alduse, Bejoy P. and Jung, Sungmoon},
  title   = {Hurricane loss estimation in wood-frame buildings using Bayesian model updating: Assessing uncertainty in fragility and reliability analyses},
  journal = {Engineering Structures},
  year    = {2017},
  volume  = {135},
  pages   = {81--94},
  doi     = {10.1016/j.engstruct.2016.12.063}
}

@article{lei2021seismic,
  author  = {Lei, Xiaoming and Sun, Limin and Xia, Ye},
  title   = {Seismic fragility assessment and maintenance management on regional bridges using bayesian multi-parameter estimation},
  journal = {Bulletin of Earthquake Engineering},
  year    = {2021},
  volume  = {19},
  pages   = {6693--6717},
  doi     = {10.1007/s10518-021-01072-6}
}

@article{nofal2020minimal,
  author  = {Nofal, Omar M. and {van de Lindt}, John W.},
  title   = {Minimal Building Flood Fragility and Loss Function Portfolio for Resilience Analysis at the Community Level},
  journal = {Water},
  year    = {2020},
  volume  = {12},
  number  = {8},
  pages   = {2277},
  doi     = {10.3390/w12082277}
}

@inproceedings{reis2022methodology,
  author    = {dos Reis, Fernando Bereta and Royer, Patrick and Chalishazar, Vishvas Hiren and Davis, Sarah and Elizondo, Marcelo and Dagle, Jeffery and Nassif, Alexandre and Sadikovic, Andrija and Ntakou, Elli and Soto, Olga and Bahramirad, Shay},
  title     = {Methodology to Calibrate Fragility Curves Using Limited Real-World Data},
  booktitle = {2022 IEEE Power \& Energy Society General Meeting (PESGM)},
  pages     = {1--5},
  year      = {2022},
  doi       = {10.1109/PESGM48719.2022.9916809}
}

@article{harirchian2024utilizing,
  author  = {Harirchian, Ehsan and Aghakouchaki Hosseini, Seyed Ehsan and Novelli, Viviana and Lahmer, Tom and Rasulzade, Shahla},
  title   = {Utilizing advanced machine learning approaches to assess the seismic fragility of non-engineered masonry structures},
  journal = {Results in Engineering},
  year    = {2024},
  volume  = {21},
  pages   = {101750},
  doi     = {10.1016/j.rineng.2024.101750}
}

@article{figueira2025virtual,
  author  = {Figueira, S. A. and Amini, M. and Cox, D. T. and Barbosa, A. R.},
  title   = {Methodology for Virtual Damage Assessment and First-Floor Elevation Estimation: Application to Fort Myers Beach, Florida and Hurricane Ian (2022)},
  journal = {Natural Hazards Review},
  year    = {2025},
  volume  = {26},
  number  = {2},
  pages   = {04025012},
  doi     = {10.1061/NHREFO.NHENG-2310}
}

@misc{kaiser2023adcirc,
  author       = {Kaiser, Chris and Dawson, Clint N. and Nikidis, Elias and Fleming, Joannes G.},
  title        = {ADCIRC/SWAN Hindcasts for Historical Storms 2003--2023},
  howpublished = {DesignSafe-CI},
  year         = {2023},
  doi          = {10.17603/DS2-B5GH-CE94}
}

@inproceedings{saeednejad2025bridging,
  author    = {Saeednejad, Narges and Padgett, Jamie Ellen},
  title     = {Bridging Data Gaps in Fragility Modeling for Coastal Infrastructure Resilience},
  booktitle = {Proceedings of the 14th International Conference on Structural Safety and Reliability (ICOSSAR'25)},
  address   = {Los Angeles, California, USA},
  year      = {2025},
  note      = {1--6 June 2025}
}

@article{baker2015efficient,
  author  = {Baker, Jack W.},
  title   = {Efficient Analytical Fragility Function Fitting Using Dynamic Structural Analysis},
  journal = {Earthquake Spectra},
  year    = {2015},
  volume  = {31},
  number  = {1},
  pages   = {579--599},
  doi     = {10.1193/021113EQS025M}
}

@article{vanDeLindt2009performance,
  author  = {van de Lindt, John W. and Dao, Thang N.},
  title   = {Performance-Based Wind Engineering for Wood-Frame Buildings},
  journal = {Journal of Structural Engineering},
  year    = {2009},
  volume  = {135},
  number  = {2},
  pages   = {169--177},
  doi     = {10.1061/(ASCE)0733-9445(2009)135:2(169)}
}

@article{karamlou2015computation,
  author  = {Karamlou, Aman and Bocchini, Paolo},
  title   = {Computation of Bridge Seismic Fragility by Large-Scale Simulation for Probabilistic Resilience Analysis},
  journal = {Earthquake Engineering \& Structural Dynamics},
  year    = {2015},
  volume  = {44},
  number  = {12},
  pages   = {1959--1978},
  doi     = {10.1002/eqe.2567}
}

@article{bandini2022seismic,
  author  = {Conde Bandini, Pedro Alexandre and Padgett, Jamie Ellen and Paultre, Patrick and Siqueira, Gustavo Henrique},
  title   = {Seismic Fragility of Bridges: An Approach Coupling Multiple-Stripe Analysis and Gaussian Mixture for Multicomponent Structures},
  journal = {Earthquake Spectra},
  year    = {2022},
  volume  = {38},
  number  = {1},
  pages   = {254--282},
  doi     = {10.1177/87552930211036164}
}

@article{rincon2024fragility,
  author  = {Rincon, Raul and Padgett, Jamie Ellen},
  title   = {Fragility Modeling Practices and Their Implications on Risk and Resilience Analysis: From the Structure to the Network Scale},
  journal = {Earthquake Spectra},
  year    = {2024},
  volume  = {40},
  number  = {1},
  pages   = {647--673},
  doi     = {10.1177/87552930231219220}
}

@manual{hazus2024eq,
  author       = {{Federal Emergency Management Agency}},
  title        = {Hazus Earthquake Model Technical Manual},
  organization = {Federal Emergency Management Agency},
  year         = {2024}
}

@article{chen2025secondgen,
  author  = {Chen, Shanshan and Xie, Yazhou and Wu, Chenhao and Burton, Henry V. and Padgett, Jamie E. and Zsarn{\'o}czay, {\'A}d{\'a}m},
  title   = {Second-generation Component- and System-level Seismic Fragility Models for Reinforced Concrete Bridges in California},
  journal = {Earthquake Spectra},
  year    = {2025},
  volume  = {41},
  number  = {4},
  pages   = {3234--3253},
  doi     = {10.1177/87552930251343634}
}

@article{lee2025systematic,
  author  = {Lee, Jeonghyun and Lochhead, Meredith and Zhong, Kuanshi and Deierlein, Gregory G.},
  title   = {Systematic Training and Validation of Parameterized Probabilistic Learning on Manifolds Surrogate Model for Seismic Performance Assessment of Highway Bridges},
  journal = {Earthquake Engineering \& Structural Dynamics},
  year    = {2025},
  volume  = {54},
  number  = {15},
  pages   = {3726--3745},
  doi     = {10.1002/eqe.70052}
}

@article{cremen2019benchmarking,
  author  = {Cremen, Gemma and Baker, Jack W.},
  title   = {A Methodology for Evaluating Component-Level Loss Predictions of the FEMA P-58 Seismic Performance Assessment Methodology},
  journal = {Earthquake Spectra},
  year    = {2019},
  volume  = {35},
  number  = {4},
  pages   = {1937--1959}
}

@article{wing2020new,
  author  = {Wing, Oliver E. J. and Pinter, Nicholas and Bates, Paul D. and Kousky, Carolyn},
  title   = {New insights into US flood vulnerability revealed from flood insurance big data},
  journal = {Nature Communications},
  year    = {2020},
  volume  = {11},
  pages   = {1444},
  doi     = {10.1038/s41467-020-15264-2}
}

@article{forte2025flash,
  author  = {Forte, Giuseppina and others},
  title   = {Flash flood impacts and vulnerability mapping at catchment scale: developing fragility curves from post-event damage data},
  journal = {Engineering Geology},
  year    = {2025}
}

@article{hoffman2014nuts,
  author  = {Hoffman, Matthew D. and Gelman, Andrew},
  title   = {The No-U-Turn Sampler: Adaptively Setting Path Lengths in Hamiltonian Monte Carlo},
  journal = {Journal of Machine Learning Research},
  year    = {2014},
  volume  = {15},
  number  = {47},
  pages   = {1593--1623}
}

@phdthesis{mangalathu2017thesis,
  author  = {Mangalathu, Sujith},
  title   = {Performance Based Grouping and Fragility Analysis of Box-Girder Bridges in California},
  school  = {Georgia Institute of Technology},
  address = {Atlanta, GA},
  year    = {2017}
}

@article{douglass2006impact,
  author  = {Douglass, Scott L. and Hughes, Steven A. and Rogers, Spencer and Chen, Qin},
  title   = {The Impact of Hurricane {Ivan} on the Coastal Roads of {Florida} and {Alabama}: A Preliminary Report},
  journal = {Report to the Coastal Transportation Engineering Research and Education Center, University of South Alabama},
  year    = {2006}
}

@article{bradner2011experimental,
  author  = {Bradner, Christopher and Schumacher, Thomas and Cox, Daniel and Higgins, Christopher},
  title   = {Experimental Setup for a Large-Scale Bridge Superstructure Model Subjected to Waves},
  journal = {Journal of Waterway, Port, Coastal, and Ocean Engineering},
  year    = {2011},
  volume  = {137},
  number  = {1},
  pages   = {3--11},
  doi     = {10.1061/(ASCE)WW.1943-5460.0000059}
}

@techreport{aashto2008guide,
  author      = {{AASHTO}},
  title       = {Guide Specifications for Bridges Vulnerable to Coastal Storms},
  institution = {American Association of State Highway and Transportation Officials},
  address     = {Washington, D.C.},
  year        = {2008}
}

@article{gretton2012kernel,
  author  = {Gretton, Arthur and Borgwardt, Karsten M. and Rasch, Malte J. and Sch\"olkopf, Bernhard and Smola, Alexander},
  title   = {A Kernel Two-Sample Test},
  journal = {Journal of Machine Learning Research},
  year    = {2012},
  volume  = {13},
  pages   = {723--773}
}

\end{document}